\documentclass{article}

   \usepackage[nonatbib,preprint]{neurips_2026}
\usepackage[numbers]{natbib}
\usepackage[utf8]{inputenc} %
\usepackage[T1]{fontenc}    %
\usepackage{hyperref}       %
\usepackage{url}            %
\usepackage{booktabs}       %
\usepackage{amsfonts}       %
\usepackage{nicefrac}       %
\usepackage{microtype}      %
\usepackage[dvipsnames]{xcolor}         %

\usepackage{amsmath}
\definecolor{cvprblue}{rgb}{0.21,0.49,0.74}
\usepackage{multirow}
\usepackage{wrapfig}
\usepackage{amssymb} 
\usepackage{pifont}
\usepackage{graphicx}
\usepackage{subcaption}
\usepackage{hyperref}
\usepackage{caption}

\newcommand{\cmark}{\textcolor{black}{\ding{51}}}%
\newcommand{\xmark}{\textcolor{black}{\ding{55}}}%
\definecolor{cerulean}{RGB}{42,82,190}
\newcommand{\rownumber}[1]{\textcolor{cerulean}{#1}}

\newcommand{\ourmethod}{LOCI}

\definecolor{locatorcol}{RGB}{55, 73, 107}
\definecolor{criticcol}{RGB}{125, 86, 47}

\usepackage{tcolorbox}
\tcbuselibrary{breakable, skins, listings}
\newtcblisting{systemlocatorprompt}[1][]{
  colback=blue!5,
  colframe=blue!40,
  coltitle=white,
  title=\textbf{Locator System Prompt},
  fonttitle=\large,
  sharp corners=south,
  breakable,
  enhanced,
  listing only,     %
  listing options={
    basicstyle=\small\ttfamily,
    breaklines=true,       %
    columns=fullflexible,  %
    keepspaces=true,       %
    showstringspaces=false,%
    literate={~} {$\sim$}{1}, %
  },
  #1
}
\newtcblisting{locatorprompt}[1][]{
  colback=blue!5,
  colframe=blue!40,
  coltitle=white,
  title=\textbf{Locator Agent Prompt},
  fonttitle=\large,
  sharp corners=south,
  breakable,
  enhanced,
  listing only,     %
  listing options={
    basicstyle=\small\ttfamily,
    breaklines=true,       %
    columns=fullflexible,  %
    keepspaces=true,       %
    showstringspaces=false,%
    literate={~} {$\sim$}{1}, %
  },
  #1
}
\newtcblisting{systemcriticprompt}[1][]{
  colback=red!5,
  colframe=red!40,
  coltitle=white,
  title=\textbf{Critic System Prompt},
  fonttitle=\large,
  sharp corners=south,
  breakable,
  enhanced,
  listing only,     %
  listing options={
    basicstyle=\small\ttfamily,
    breaklines=true,       %
    columns=fullflexible,  %
    keepspaces=true,       %
    showstringspaces=false,%
    literate={~} {$\sim$}{1}, %
  },
  #1
}
\newtcblisting{criticprompt}[1][]{
  colback=red!5,
  colframe=red!40,
  coltitle=white,
  title=\textbf{Critic Agent Prompt},
  fonttitle=\large,
  sharp corners=south,
  breakable,
  enhanced,
  listing only,     %
  listing options={
    basicstyle=\small\ttfamily,
    breaklines=true,       %
    columns=fullflexible,  %
    keepspaces=true,       %
    showstringspaces=false,%
    literate={~} {$\sim$}{1}, %
  },
  #1
}
\newtcblisting{systemmultilocatorprompt}[1][]{
  colback=NavyBlue!5,
  colframe=NavyBlue!40,
  coltitle=white,
  title=\textbf{Multi-Locator System Prompt},
  fonttitle=\large,
  sharp corners=south,
  breakable,
  enhanced,
  listing only,     %
  listing options={
    basicstyle=\small\ttfamily,
    breaklines=true,       %
    columns=fullflexible,  %
    keepspaces=true,       %
    showstringspaces=false,%
    literate={~} {$\sim$}{1}, %
  },
  #1
}
\newtcblisting{multilocatorprompt}[1][]{
  colback=NavyBlue!5,
  colframe=NavyBlue!40,
  coltitle=white,
  title=\textbf{Multi-Locator Agent Prompt},
  fonttitle=\large,
  sharp corners=south,
  breakable,
  enhanced,
  listing only,     %
  listing options={
    basicstyle=\small\ttfamily,
    breaklines=true,       %
    columns=fullflexible,  %
    keepspaces=true,       %
    showstringspaces=false,%
    literate={~} {$\sim$}{1}, %
  },
  #1
}

\newtcblisting{multisystemcriticprompt}[1][]{
  colback=purple!5,
  colframe=purple!40,
  coltitle=white,
  title=\textbf{Critic System Prompt (for Locator Ensembling)},
  fonttitle=\large,
  sharp corners=south,
  breakable,
  enhanced,
  listing only,     %
  listing options={
    basicstyle=\small\ttfamily,
    breaklines=true,       %
    columns=fullflexible,  %
    keepspaces=true,       %
    showstringspaces=false,%
    literate={~} {$\sim$}{1}, %
  },
  #1
}
\newtcblisting{multicriticprompt}[1][]{
  colback=purple!5,
  colframe=purple!40,
  coltitle=white,
  title=\textbf{Critic Agent Prompt (for Locator Ensembling)},
  fonttitle=\large,
  sharp corners=south,
  breakable,
  enhanced,
  listing only,     %
  listing options={
    basicstyle=\small\ttfamily,
    breaklines=true,       %
    columns=fullflexible,  %
    keepspaces=true,       %
    showstringspaces=false,%
    literate={~} {$\sim$}{1}, %
  },
  #1
}

\newtcblisting{systemfinalagentprompt}[1][]{
  colback=ForestGreen!5,
  colframe=ForestGreen!40,
  coltitle=white,
  title=\textbf{Final Answering Agent System Prompt (for Locator Ensembling)},
  fonttitle=\large,
  sharp corners=south,
  breakable,
  enhanced,
  listing only,     %
  listing options={
    basicstyle=\small\ttfamily,
    breaklines=true,       %
    columns=fullflexible,  %
    keepspaces=true,       %
    showstringspaces=false,%
    literate={~} {$\sim$}{1}, %
  },
  #1
}
\newtcblisting{finalagentprompt}[1][]{
  colback=ForestGreen!5,
  colframe=ForestGreen!40,
  coltitle=white,
  title=\textbf{Final Answering Agent Prompt (for Locator Ensembling)},
  fonttitle=\large,
  sharp corners=south,
  breakable,
  enhanced,
  listing only,     %
  listing options={
    basicstyle=\small\ttfamily,
    breaklines=true,       %
    columns=fullflexible,  %
    keepspaces=true,       %
    showstringspaces=false,%
    literate={~} {$\sim$}{1}, %
  },
  #1
}

\title{LOCI: A Locator-Critic with Refinement Loop }

\author{%
  Walid Bousselham \\
  \And
  Mathilde Caron \\
  \And
  Arsha Nagrani \\
  \And
  Cordelia Schmid \\
  \And
  \\
  Google DeepMind\\
}

\begin{document}

\maketitle

\begin{abstract}
  Vision-Language Models (VLMs) still struggle on tasks requiring 
complex visual understanding. 
We argue that the core issue is not high-level reasoning, but instead failing to locate critical details in the image.
Due to this shortcoming, VLMs generate often plausible but incorrect reasoning based on flawed perceptual grounding. 
To address this, we propose Locator-Critic (LOCI), a training-free framework that decouples visual search from evidence verification. 
LOCI employs a Locator agent to propose candidate visual evidence and a separate Critic agent to evaluate its relevance and sufficiency. 
These agents engage in an iterative refinement loop, progressively improving the evidence until it is adequate to answer the given question. 
This decoupled, self-correcting process yields substantial performance gains, achieving state-of-the-art results on multiple complex visual benchmarks. 
LOCI improves accuracy for both open-weight models like Qwen3-VL (+12.1 on V*, +5.8 on HR-Bench and +11.2 on VisualProbe-Hard) and proprietary models like Gemini 2.5 Pro (+8.9 on V*, +4.3 on HR-Bench, +4.8 on VisualProbe-Hard). 
\end{abstract}

\section{Introduction}
\label{sec:intro}

Modern Vision-Language Models~\cite{comanici2025gemini, jaech2024openai, ZhangOpenAIOS} produce sophisticated reasoning traces and thought processes.
However, their performance remains limited on tasks demanding detailed and complex visual analysis. 
We argue this is not a failure of the reasoning process itself.
Instead, the main bottleneck is the ability to reliably \textit{locate} the specific visual cues necessary to answer a question in complex scenes.
We observe that when a VLM focuses on an irrelevant or ambiguous region, its reasoning process generates a narrative based on this flawed visual, leading to an incorrect answer.

We argue that this issue arises because the model must both propose and verify visual evidence in a single inference call.
Once the model identifies a visual region, it is heavily biased toward confirming its own hypothesis rather than rigorously challenging it.
As a result, identifying the wrong visual region often leads to a wrong answer.

To confirm our main hypothesis we carefully design an oracle experiment (see Tab.~\ref{tab:oracle_results}), where we manually provide the ground-truth visual regions to the model.
When provided with the correct visual cues, the model is capable of answering almost all the questions of V* correctly with a valid reasoning process, confirming our main hypothesis.
The performance is not bounded by reasoning ability, but primarily by the ability to locate relevant visual information.
To mitigate this issue, we propose separating the reasoning process into distinct localization  and verification phases.

\begin{figure*}[t]
    \vspace{-1em}
    \centering
    \includegraphics[width=\linewidth]{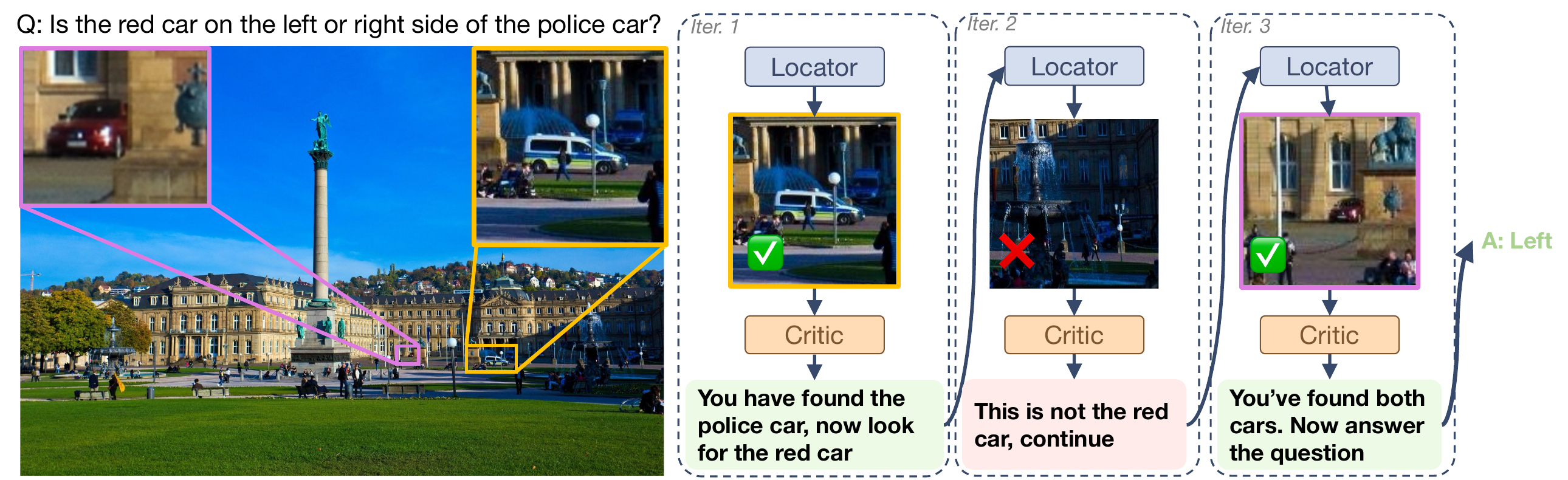}
    \caption{
    \textbf{LOCI approach.}
To answer the question, our method employs a Locator to find relevant objects and a Critic to provide feedback. It first finds the police car (Iter. 1), then, after a failed attempt (Iter. 2), successfully locates the red car (Iter. 3). With both objects grounded, it correctly infers their spatial relationship to provide the answer.
    }
    \label{fig:teaser}
    \vspace{-2em}
\end{figure*}

We introduce the LOCI approach (Locator-Critic with Refinement), a training-free agentic pipeline that decouples (i) finding visual evidence and (ii) verification, using two specialized agents.
These agents are the same underlying model, differentiated only by their prompts and inputs.
The \textbf{Locator} proposes visual evidence by extracting relevant crops from the image.
The \textbf{Critic} independently evaluates whether the proposed crops are sufficient to answer the question, judging only the visual evidence without access to the Locator's textual reasoning. Importantly, the Critic only accesses the visual evidence and is not provided with the Locator's textual reasoning. As we demonstrate in Table~\ref{tab:locator_comparison}, this separation is crucial to prevent the Locator from propagating its own errors and influencing the Critic's assessment.
Their interaction forms a self-correcting loop. 
When the Critic rejects the evidence as insufficient, it provides targeted feedback that guides the Locator to search again, as illustrated in Figure~\ref{fig:teaser}.

We find that the \textbf{LOCI} framework is particularly suited for benchmarks with complex and 
detailed images,
such as V*~\cite{wu2024v}, HR-Bench~\cite{wang2025divide}, and VisualProbe~\cite{lai2025mini}.
On these benchmarks, our training-free Locator-Critic framework achieves state-of-the-art performance, surpassing prior methods, including recent reinforcement learning-based approaches~\cite{lai2025mini, zheng2025deepeyes, su2025pixel, zhang2025chain}. 
We also show that \textbf{LOCI} is flexible, demonstrating consistent improvements across both open-weight models (Qwen3-VL: +12.1/5.8/11.2 points on V*/HR-Bench/VisalProbe) and proprietary models (Gemini2.5-Pro: +8.9/4.3/4.8 points on V*/HR-Bench/VisualProbe).
\noindent Overall, our contributions can be summarized as
:
\begin{itemize}
\vspace{-1em}
\item We propose \textbf{LOCI}, a training-free Locator-Critic framework that improves VLM reliability by decoupling visual evidence finding from evidence verification.
\item We significantly outperform the SOTA on multiple benchmarks for Qwen3-VL~\cite{qwen3vl} and Gemini2.5-Pro~\cite{comanici2025gemini}, with consistent improvements over direct prompting and prior methods.
\item Extensive ablation studies demonstrate the critical role of decoupled verification and the effectiveness of iterative refinement. We also demonstrate scalability through parallel multi-locator ensembling,
\end{itemize}
\vspace{-1.em}

\section{Related Works}
\label{sec:related_works}

\subsection{Dynamic Visual Computation and Tool-Use}
Recently, tool-augmented MLLMs \cite{gupta2023visual, yang2023mm, team2025kimi, zhang2025thyme, lai2025mini, zheng2025deepeyes, su2025pixel} have shown strong agentic abilities in complex QA tasks by leveraging
a broad set of tools (e.g., web browsing, code execution, retrieval). Specifically for image manipulation, Thyme~\cite{zhang2025thyme} uses code generation with a diverse set of operations like cropping, rotation, and contrast enhancement, activated in the model via SFT and RL stages.  Mini-O3~\cite{lai2025mini} similarly handles complex queries through on-the-fly image manipulation. 
While powerful, these approaches often require significant training overhead. 
Our method, in contrast, is entirely training-free and generalizes across both proprietary and open-source models.

\begin{figure*}[t]
    \vspace{-1em}
    \centering
    \includegraphics[width=1.\linewidth]{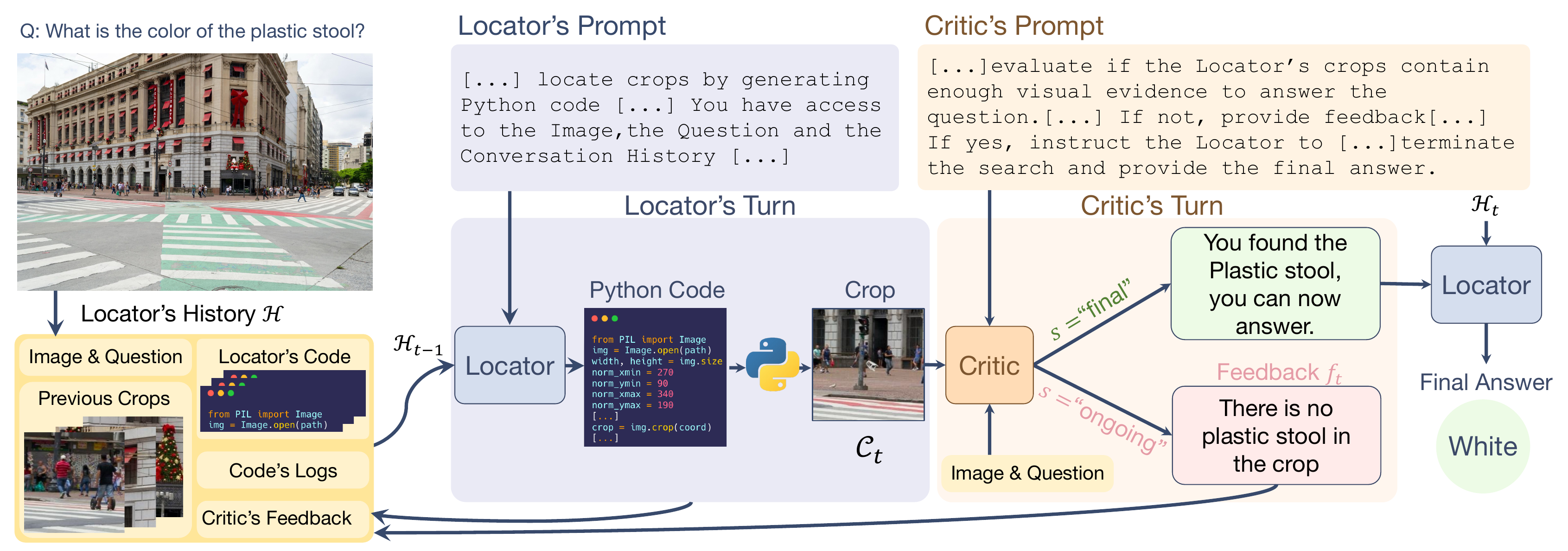}
    \caption{
    \textbf{Overview of \ourmethod~.} 
    A Locator agent generates Python code to create image crops, which are then evaluated by a Critic agent. The Critic's feedback guides the Locator's following attempts until sufficient visual evidence is found to answer the question.
    }
    \label{fig:method_fig}
    \vspace{-1em}
\end{figure*}

Similar to our work, DeepEyes\cite{zheng2025deepeyes}, Chain-of-Focus\cite{zhang2025chain} and Pixel Reasoner\cite{su2025pixel}, all aim
to equip VLMs with zoom-in and region-of-interest selection, enabling active perception over images.
Similarly, DyFo~\cite{li2025dyfo} employs a training-free Monte Carlo Tree Search to guide a visual expert's focus, while Token-Efficient VLM~\cite{jiang2024teva} and Griffon-G~\cite{zhan2024griffon} use complex, multi-stage training pipelines to select high-resolution patches or generate textual coordinates.
Unlike these works, we decouple visual detection from verification through a dedicated Critic agent that evaluates the quality of generated crops and provides targeted feedback for iterative refinement.

\subsection{Critique and Verification in MLLMs}
Multiple works try and improve the reliability of models by incorporating an explicit `critic' or `verifier' module. 
The simplest form of this is self-correction, where a single model critiques its own output. 
This has been explored in both text-only (eg.\ SELF-REFINE~\cite{madaan2023self}) and multimodal domains (LLaVA-Critic-R1~\cite{wang2025llava}).  While straightforward to implement, these approaches can suffer from "cognitive fixation," where the model fails to identify its own initial perceptual or reasoning errors because the critic and generator share the same underlying biases.

To overcome this limitation, recent work has moved towards a decoupled "actor-critic" paradigm, where an independent model verifies the output of the primary reasoning model. 
For example, Critic-V~\cite{zhang2025critic} uses a critic to provide natural language feedback on a reasoning model's thought process. MMC~\cite{liu2025mmc} automatically constructs a critique dataset by using Monte Carlo Tree Search to compare correct and incorrect reasoning paths, while OmniVerifier~\cite{zhang2025generative} trains a universal verifier to assess the alignment of generated visual outcomes with textual prompts.

Our LOCI framework builds upon and extends this actor-critic paradigm with two crucial differences. 
First, unlike general-purpose frameworks like LLM-as-a-Judge~\cite{zheng2023judging} that evaluate a final answer, our Critic is deployed at an intermediate reasoning stage. 
Its specific role is to evaluate the quality of the visual evidence (the proposed crop) for answering the given query, not the final textual response. 
Second, to ensure an unbiased assessment, our Critic is intentionally decoupled from the Locator's reasoning process. 
It receives only the candidate crop and the original question, forcing it to judge the visual evidence on its own.

\section{Method}
\label{sec:method}

\label{sec:locator_critic}

We introduce the Locator-Critic framework (\ourmethod~) for effective visual exploration and verification. Our framework decouples visual search from verification.
We also demonstrate how to scale this framework with ensembling multiple locators working simultaneously on the same task.

\subsection{Locator-Critic Framework}
Our multi-agent pipeline consists of two agents working collaboratively to solve the visual question answering task (see Figure~\ref{fig:method_fig}). A \textbf{Locator} visually explores the image by producing crops to gather visual evidence sufficient for answering the question (purple area of Fig.~\ref{fig:method_fig}), while a \textbf{Critic} analyzes the visual evidence produced by the Locator and decides whether sufficient reliable evidence has been found to properly answer the question (orange area of Fig.~\ref{fig:method_fig}) .

\paragraph{The Locator Agent.}
The Locator's role is to search for and extract crops containing visual evidence needed to answer the question. The Locator is a VLM, which we prompt specifically (see \textit{"Locator’s Prompt
"} in Fig.~\ref{fig:method_fig} for a short version and Appendix~\ref{sec:supp_prompts_loci} for the full prompt) to generate both bounding box coordinates and Python code to create crops, following prior works~\cite{li2025torl, zhang2025thyme, alrashedy2024generating}.
The generated code is executed and the model receives the outputs (e.g. logs, error messages, etc.) and any image saved (e.g. a crop).
This approach enables flexible and adaptive visual detection with iterative refinement based on Critic feedback for example.

Formally, at each turn $t$ ($ 1 \le t \le T_{\max}$), the Locator receives as input a cumulative conversation history $\mathcal{H}_{t-1}$ illustrated as a yellow box in the bottom left of Fig.~\ref{fig:method_fig}.
At initialization (i.e. $t=1$), the history $\mathcal{H}_{0}$ contains solely the question and the source image.
At each turn, the task of the Locator is to output a set of maximum $m$ crops $\mathcal{C}_t = \{C_1, \ldots, C_m\}$ containing visual evidence needed to answer the question.
Importantly the Locator finally answers the question $Q$ only when explicitly prompted to do so, \textit{i.e.} at the final turn.

The conversation history $\mathcal{H}_{t-1}$ input to the Locator at turn $t$ is composed of different artifacts from all preceding turns ($\le t - 1$), which include: (1) the complete execution trace of the Locator (python code input, set of crops output, and code execution logs), and (2) targeted feedback provided by the Critic (see details in next paragraph).
Leveraging this comprehensive history, the Locator can refine its previous output by modifying its previous own code. For instance, it can debug its execution using past logs or adjust bounding box coordinates in its Python script in response to the Critic's feedback.

\paragraph{The Critic Agent.}
The Critic is a VLM that, at each turn $t$, receives as input the question, the original image and the Locator crops $\mathcal{C}_t$. Its task is to evaluate whether the crops provide sufficient visual evidence to answer the question (see \textit{"Critic's Prompt"} in Fig.~\ref{fig:method_fig} for a short version and Appendix~\ref{sec:supp_prompt_critic} for the full prompt). The Critic outputs a binary decision $s \in \{\text{ongoing}, \text{final}\}$ indicating whether additional search is needed.
When additional search is needed ($s = \text{ongoing}$), the Critic provides targeted feedback $f_t$ to guide the Locator toward more relevant visual regions—for example, in Fig.~\ref{fig:method_fig} requesting another crop of the plastic stool.
This feedback $f_t$ is appended to the conversation history $\mathcal{H}_{t}$ and will be used by the Locator in the following turns.
When sufficient evidence has been gathered or when the maximal number of turns has been achieved, the Critic prompts the Locator to answer the question based on the collected visual evidences.

\paragraph{Limiting the Critic access to the Locator reasoning.}
We find that a critical design principle is to provide the Critic only the crops as visual evidence, not the Locator's textual reasoning. 
We show (see Tab.~\ref{tab:critic_ablation} \& Fig.~\ref{sup:fig:critic_text}) that otherwise the Critic can be deceived by the Locator's explanations into approving insufficient visual evidence.
This decoupling mitigates hallucination and ensures genuine visual verification.

\paragraph{Off-the-Shelf Detectors.}
While our core framework uses VLM-based code generation for flexible crop extraction, one could alternatively employ off-the-shelf open-vocabulary object detectors (e.g., OWLv2~\cite{minderer23neurips_owlv2}) to generate fixed object-centric crops. However, such detectors produce non-refinable outputs, limiting the Critic's ability to guide iterative improvement. Our experiments (see Tab.~\ref{tab:locator_comparison}) demonstrate that using an off-the-shelf detection results in comparable results as the first turn, but cannot be improved further. This is interesting as it confirms (i) the importance of visual selection and (ii) the importance of the iterative refinement with a critic, which is shown to significantly improve the performance.

\subsection{Multi-Locator Ensembling}
\label{sec:multi_locator}
We further scale the framework by deploying multiple locators in parallel. Due to the non-deterministic nature of VLMs, each of the agents outputs a different set of crops for the same input image and question.  This strategy increases exploration diversity and robustness, improving the chances of finding the correct visual evidence.

\paragraph{Ensembling Multiple Locators.}
We find that a single locator may become stuck exploring the wrong region and, in practice, may not recover from an initial incorrect hypothesis (see qualitative examples in Sec.~\ref{fig:qualitative_example}).
Since our framework decouples visual search from verification, we can deploy multiple locators ($k$ instances) that search simultaneously. This provides effective mitigation for failure cases where one locator gets stuck, while enabling ensemble scaling of visual visual detection capacity, the benefits of which are shown in Table~\ref{tab:ablation_locators}.
When ensembling multiple locators, we rely on $k$ locators $\mathcal{L} = \{L_1, \ldots, L_k\}$ in parallel per iteration.
Using non-zero temperature for code generation ensures that different locator instances produce diverse exploration strategies, even when given the same input.
Each locator $L_i$ independently generates crops and maintains its own conversation history $\mathcal{H}_t^{(i)}$, allowing it to explore different spatial regions simultaneously.

We then employ a centralized Critic that evaluates all crops from all locators collectively as $\mathcal{C}^{(\text{all})}_{t} = \bigcup_{i=1}^{k} \mathcal{C}_t^{(i)}$. 
At each iteration, the Critic determines if the combined evidence is sufficient. If it is not, the Critic generates a unified feedback message that is broadcast to all locators to guide their next search iteration. This feedback is typically a simple directive, such as indicating that the current crops are insufficient or suggesting an unexplored region of the image to focus on.
Each locator appends this feedback to its own conversational history, and the process continues to iteration $t+1$.
When sufficient evidence has been gathered, the Critic selects the best crops $\mathcal{C}_{\text{selected}}  \subseteq \mathcal{C}^{(\text{all})}_{t}$ and proceeds to final answering. Eventually, the final answering agent receives the selected crops $\mathcal{C}_{\text{selected}}$ along with the question and full image to generate the answer.

\section{Experiments}
\label{sec:experiments}
\subsection{Experimental Setup}
\label{sec:experiments:setup}

\paragraph{Datasets.} 

We evaluate our method on three VQA benchmarks that challenge modern VLMs with complex questions where there is still substantial potential for improvement.
\textbf{V*}~\cite{wu2024v} consists of $191$ complex visual reasoning questions. The objects relevant to the questions are annotated with ground truth bounding boxes. This enables  oracle experiments to quantify the performance with perfect localization.  \textbf{HR-Bench}~\cite{wang2025divide}-  composed of 177 images and 800 questions- focuses on high-resolution image understanding with a focus on object-centric questions. 
\textbf{VisualProbe}~\cite{lai2025mini} consists of 106 samples featuring visually complex scenes with many interacting objects, paired with open-ended questions.
We report results on the test set of the 3 benchmarks.

\paragraph{Implementation Details.}
Our core framework uses a single Locator paired with a Critic.
By default, we use a maximum of $T_{\max}=10$ turns and limit the number of crops per turn to $m=4$ (see ablation in Appendix~\ref{sec:supp_ablation_crops}).
We use a single locator ($k=1$), unless stated otherwise.
We evaluate our framework using two SOTA models: (i) Gemini2.5-Pro~\cite{comanici2025gemini}, a proprietary model with thinking, and (ii) Qwen3-VL-235b-a22b-Thinking~\cite{qwen3vl}, an open-weight model referred to as Qwen3-VL.

\subsection{Ablation Study}
\label{sec:experiments:ablations}
In this section, unless specified otherwise, we conduct all of the experiments on V* benchmark using Gemini2.5-Pro. We provide additional ablation results on more datasets in Appendix~\ref{sec:supp_ablation_datasets}.

\paragraph{Oracle Experiment.}\label{sec:exp:oracle_results}

\begin{wraptable}[7]{r}{0.35\textwidth}
\vspace{-2.em}
\small
  \caption{\textbf{Oracle experiment on $V^*$.} We provide the ground-truth crops required to answer the question.
  }
  \label{tab:oracle_results}
  \centering
  \begin{tabular}{@{}lc@{}}
    \toprule
    Method & acc. \\
    \midrule
    Gemini2.5-Pro & 83.8 \\
    Gemini2.5-Pro + Oracle Crop & 98.2\\
    \bottomrule
  \end{tabular}
\end{wraptable}

To test the hypothesis that visual localization, rather than reasoning capability, is the primary bottleneck, we conduct an oracle experiment on V*~\cite{wu2024v}.
We provide the model with ground truth bounding boxes from the dataset, extracting oracle crops around objects of interest. These crops are provided together with the full image and the question to Gemini2.5-Pro.
As shown in Table~\ref{tab:oracle_results}, this oracle experiment confirms that visual localization is the primary bottleneck.
Providing the model with perfect localization information (oracle crops and their coordinates) dramatically improves accuracy from 83.0\% to 98.2\%.
A manual inspection of the remaining  1.8\% failure cases shows that they correspond to  ambiguous questions with multiple viable answers (see Appendix~\ref{sec:supp_oracle_failures} for examples).

\begin{wraptable}[13]{r}{0.55\textwidth}
\vspace{-2em}
\small
  \caption{
    \textbf{Locator \& Critic ablations.}
    We show the impact of the Locator \& Critic in the LOCI framework:
    having a Locator improves direct prompting by +5.4 (row~\rownumber{1} \textit{vs} row~\rownumber{2}).
Also, having a Critic gives a +3.5 boost (row~\rownumber{2} \textit{vs} row~\rownumber{4}).
It is crucial not to provide the Locator text as input to the Critic (row~\rownumber{3} \textit{vs} row~\rownumber{4})
  }
  \label{tab:critic_ablation}
    \setlength{\tabcolsep}{4pt}
  \centering
  \begin{tabular}{@{}p{.8em}@{} c c cc c c@{}}
    \toprule
    &  Locator && \multicolumn{2}{c}{Critic} &&  Acc. \\
    \cmidrule{2-2} \cmidrule{4-5} \cmidrule{7-7}
     & && input: Locator text & input: Locator crops && \\
      \midrule
     \rownumber{1} & \xmark && \xmark & \xmark && 83.8\\
    \rownumber{2} & \cmark && \xmark & \xmark && 89.2\\ 
   \rownumber{3}  & \cmark && \cmark & \cmark && 87.2\\ 
 \rownumber{4}    & \cmark && \xmark & \cmark && \textbf{92.7}\\ 
    \bottomrule
  \end{tabular}
\end{wraptable}

\paragraph{Locator and Critic ablations.}

Table~\ref{tab:critic_ablation} presents an ablation study of the Locator and Critic components.
First, we see that Locator only (row~\rownumber{2}) improves over direct prompting (row~\rownumber{1}) by +5.4 points.
Additionally, the Critic improves the performance further by +3.5 (row~\rownumber{2} versus row~\rownumber{4}).
Overall, our full framework LOCI achieves a +8.9 points improvement compared to direct prompting baseline.

In Table~\ref{tab:critic_ablation}, we also validate a critical design choice, and show that the Critic must receive only image crops as visual evidence, not the Locator's textual reasoning. 
Providing the Locator's reasoning to the Critic (row~\rownumber{3}) severely degrades performance, performing worse than the Locator-only baseline (row~\rownumber{2}). 
We attribute this to the Critic being misled by the Locator's incorrect textual explanations, a failure mode where it is convinced to agree with incorrect evidence after a few iterations (see Appendix~\ref{sec:supp_critic_text} for examples). 
This finding underscores that isolating the Critic with purely visual evidence is crucial for preventing the propagation of hallucinations and ensuring correct visual verification.

\paragraph{Same-Budget Baseline Comparisons.}
A natural question is whether LOCI's gains come from its specific architecture or simply from spending more inference tokens.
To answer this, we compare LOCI against several baselines at comparable or higher token budgets on V*, each representing a different strategy for spending inference-time compute.
All baselines use Gemini2.5-Pro. We test each baseline in two settings: without code execution (text-only reasoning) and with code execution (access to the same crop tools as LOCI's Locator).
Brief descriptions are provided in Appendix~\ref{sec:supp_baseline_descriptions}.

\begin{table}[t]
\small
  \caption{\textbf{Same-budget baselines on V*.} We compare LOCI against alternative inference-time strategies at comparable or higher token budgets, with and without access to code execution.}
  \label{tab:same_budget}
  \centering
  \setlength{\tabcolsep}{5pt}
  \begin{tabular}{@{}lccc@{}}
    \toprule
    Method & Code-Exec & Avg Tokens & Acc. \\
    \midrule
    Direct Prompting & \xmark & 330 & 83.8 \\
    Self-Refine & \xmark & 4,188 & 81.5 \\
    Self-Consistency@10 & \xmark & 3,632 & 84.3 \\
    \midrule
    Explicit Locate-Verify-Answer & \cmark & 4,117 & 85.9 \\
    Self-Refine & \cmark & 4,201 & 85.2 \\
    Best-of-10 & \cmark & 36,310 & 89.0 \\
    Locator only (Tab.~\ref{tab:critic_ablation}, row~\rownumber{2}) & \cmark & 9500 & 89.2 \\
    Self-Consistency@10 & \cmark & 4,281 & 89.7 \\
    \midrule
    \textbf{LOCI (ours)} & \cmark & 3,690 & \textbf{92.7} \\
    \bottomrule
  \end{tabular}
  \vspace{-2em}
\end{table}

Table~\ref{tab:same_budget} shows three patterns.
First, spending more tokens on text-only self-critique hurts: Self-Refine without code execution (81.5) drops below direct prompting (83.8), confirming that the bottleneck is perceptual rather than reasoning-based.
Second, code execution without decoupled verification plateaus around 89/90. Self-Consistency@10 with code execution (89.7), Locator-only (89.2), and Best-of-10 (89.0) all cluster at this ceiling. Best-of-10 is particularly informative: it uses nearly $10\times$ LOCI's token budget (36,310 tokens) yet still falls 3.7 points short. Even with substantially more compute, a single agent remains anchored to its initial search strategy.
Third, LOCI's decoupled architecture is the differentiator. The Explicit Locate-Verify-Answer baseline follows LOCI's exact workflow (locate, verify sufficiency, iterate) with the same tools and budget, but the verification step sees the model's own textual reasoning. The 6.8-point gap to LOCI (85.9 vs 92.7) shows that the procedure alone does not account for LOCI's gains. A model conditioned on its own reasoning cannot genuinely challenge a strategy it just committed to but decoupled verification breaks this anchoring.

\paragraph{Off-the-shelf Locator.}

\begin{figure}[t]
  \centering
  \begin{minipage}[t]{0.54\linewidth}
    \centering
    \includegraphics[width=\linewidth]{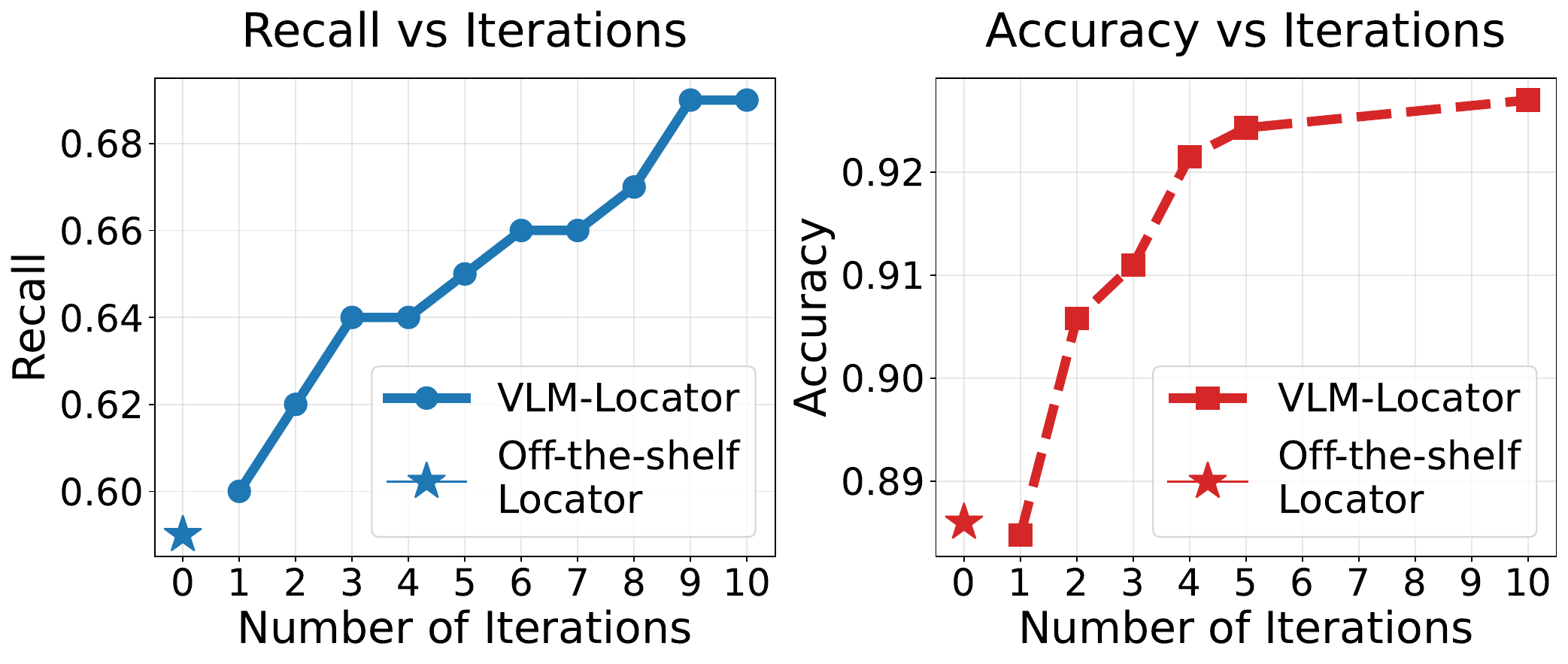}
    \caption{\textbf{Impact of Iterative Refinement on Localization and Accuracy.} We plot (left) the crops recall as a measure of Locator crops quality and (right) final VQA accuracy on the V* dataset with respect to the maximum number of turns ($T_{max}$). The strong positive correlation between the two curves demonstrates that as the iterative process improves localization (higher recall), the final task performance also increases.}
    \label{fig:recall_accuracy_iterations}
  \end{minipage}
  \hfill
  \begin{minipage}[t]{0.42\linewidth}
  \vspace{-10em}
    \centering\small
    \captionof{table}{\textbf{Different choice of Locator.} Comparison of our VLM-based Locator vs.\ an alternative off-the-shelf Locator.}
    \label{tab:locator_comparison}
    \begin{tabular}{@{}lcc@{}}
      \toprule
      & w/o Critic & with Critic \\
      \midrule
      Direct Prompting & 83.8 & - \\
      \midrule
      Off-the-shelf Locator & 88.6 & 89.3 \\
      VLM-Locator & 89.2 & 92.7 \\
      \bottomrule
    \end{tabular}

    \vspace{0.5em}

    \captionof{table}{\textbf{Computational cost.} Average token usage per sample across configurations with Qwen3-VL.}
    \label{tab:ablation_cost}
    \setlength{\tabcolsep}{2pt}
    \begin{tabular}{@{}lcc@{}}
      \toprule
      Method & avg. \# tokens & acc. \\
      \midrule
      Direct Prompting & $330$ & 83.8 \\
      Locator only & $950$ & 89.2 \\
      LOCI & $3,690$ & 92.7 \\
      \bottomrule
    \end{tabular}
  \end{minipage}
  \vspace{-2em}
\end{figure}

We compare our VLM-based Locator against an off-the-shelf object detector, OWLv2~\cite{minderer23neurips_owlv2}.
In Table~\ref{tab:locator_comparison} we test both locators types (VLM-Based vs off-the-shelf) in two configuration: one without a Critic, where the crops are passed directly to the final answering agent, and one with a Critic.
Since the off-the-shelf detector produces a fixed, non-refinable set of crops, the Critic's role in that setting is limited to a one-shot selection of the most relevant crops to pass to the final agent.

The results show that the Critic provide a slight improvement to the off-the-shelf locator +0.7 points. Conversely, the Critic yields a significant improvement for our VLM-Locator (+3.5 points).
This gap showcases a key limitation of off-the-shelf locator. 
The locator cannot refine its own predictions and the Critic's effectiveness is bounded by the recall of the initial set of crops. 
Hence, if the visual evidence necessary to answer the question is not captured in the first pass, the Critic cannot recover it.
In contrast, the iterative refinement of the VLM-Locator allows to generate better crops at each turn, leading to superior performance.

\paragraph{Evaluating the quality of the crops.}

To demonstrate that improved performance stems from higher-quality crops rather than other factors, we plot both the crop recall and the final VQA accuracy as a function of the number of Locator-Critic iterations ($T_{max}$) in Figure~\ref{fig:recall_accuracy_iterations}. We also compare the performance against an off-the-shelf Locator (OWLv2~\cite{minderer23neurips_owlv2}).

Note that evaluating the correctness of generated crops for VQA presents a challenge, as standard localization metrics are not perfectly aligned with the downstream task. 
Our VLM-Locator is instructed to produce helpful crops containing sufficient visual evidence, not necessarily perfectly tight bounding boxes. 
For instance, questions about the relative position of two objects are often best answered with a single, wider crop covering both, which would yield a low Intersection over Union (IoU) against either individual ground truth box. 
Similarly, a crop that simply zooms into a relevant region of interest is often sufficient for reasoning.
Therefore,  we adopt a lenient IoU threshold of 5\% to define a "correct" crop, which captures the practical utility of the generated regions.

Figure~\ref{fig:recall_accuracy_iterations} shows the clear benefit of iterative refinement. 
As iterations increase from $T_{max}=1$ to $T_{max}=10$, recall climbs from 0.60 to 0.69, while VQA accuracy rises from 0.89 to over 0.92. The correlation between these curves confirms that the iterative feedback loop allows the Locator to find previously missed visual evidence, and that this improved localization is the direct reason for accuracy gains. 
The metric-task misalignment also explains why the measured recall plateaus around 0.69, as many functionally optimal crops for VQA do not meet the criteria of the recall metric (see Appendix~\ref{sec:supp_metric_mismatch} for visual examples).

\paragraph{Computational Cost Analysis}
\label{sec:experiments:discussion}

In Table~\ref{tab:ablation_cost}, we analyze the computational cost of our method using Qwen3-VL.
Note that we cannot do this analysis with Gemini2.5-Pro, since it is a blackbox, and we do not have access to the tokens.
The results demonstrate a clear example of inference-time compute scaling: by increasing the number of tokens used at inference, we achieve higher accuracy without requiring any additional fine-tuning.
Our method achieves significant accuracy gains over Direct Prompting at the cost of increased computational overhead. 
Future work could explore distilling the Locator-Critic interaction into a single model to reduce this overhead while retaining the benefits of decoupled verification.

\begin{figure}[t]
\vspace{-5em}
\centering
	\begin{subfigure}[b]{1.\textwidth}
		\centering
        \includegraphics[width=\linewidth]{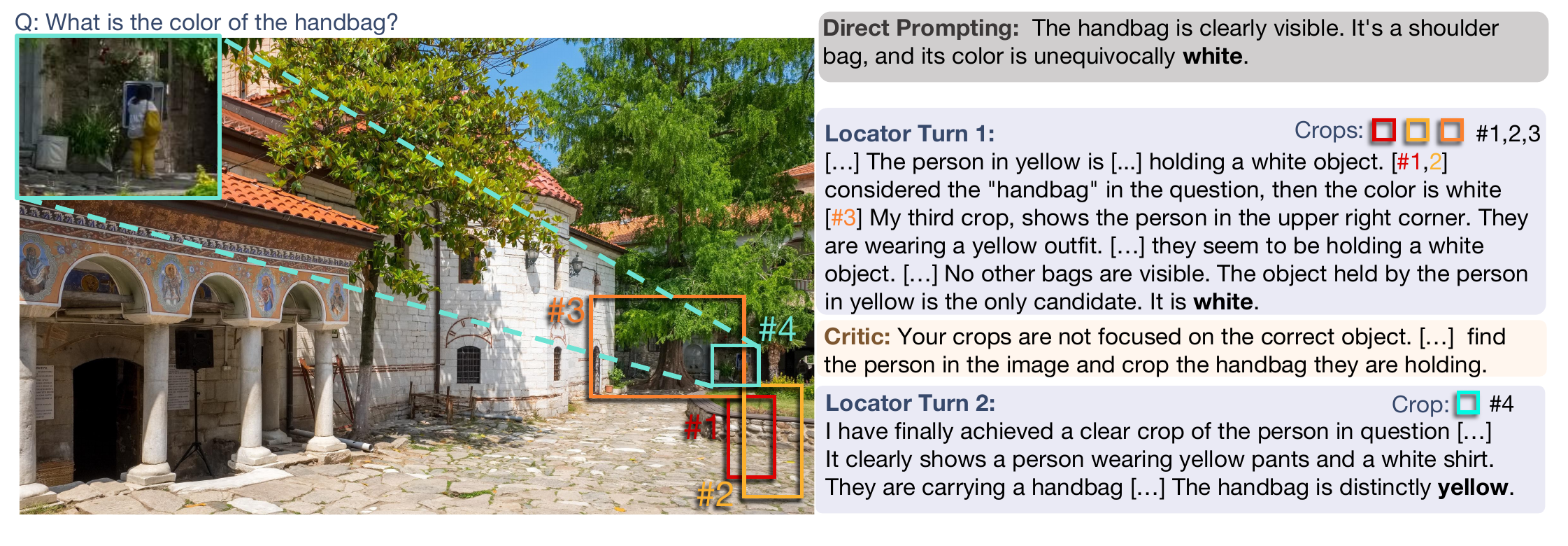}
        \vspace{-2.5em}
		\caption{Direct prompting versus LOCI}
		\label{subfig:single_vs_direct}
	\end{subfigure}
	\hfill %
	\begin{subfigure}[b]{1.\textwidth}
		\centering
        \includegraphics[width=\linewidth]{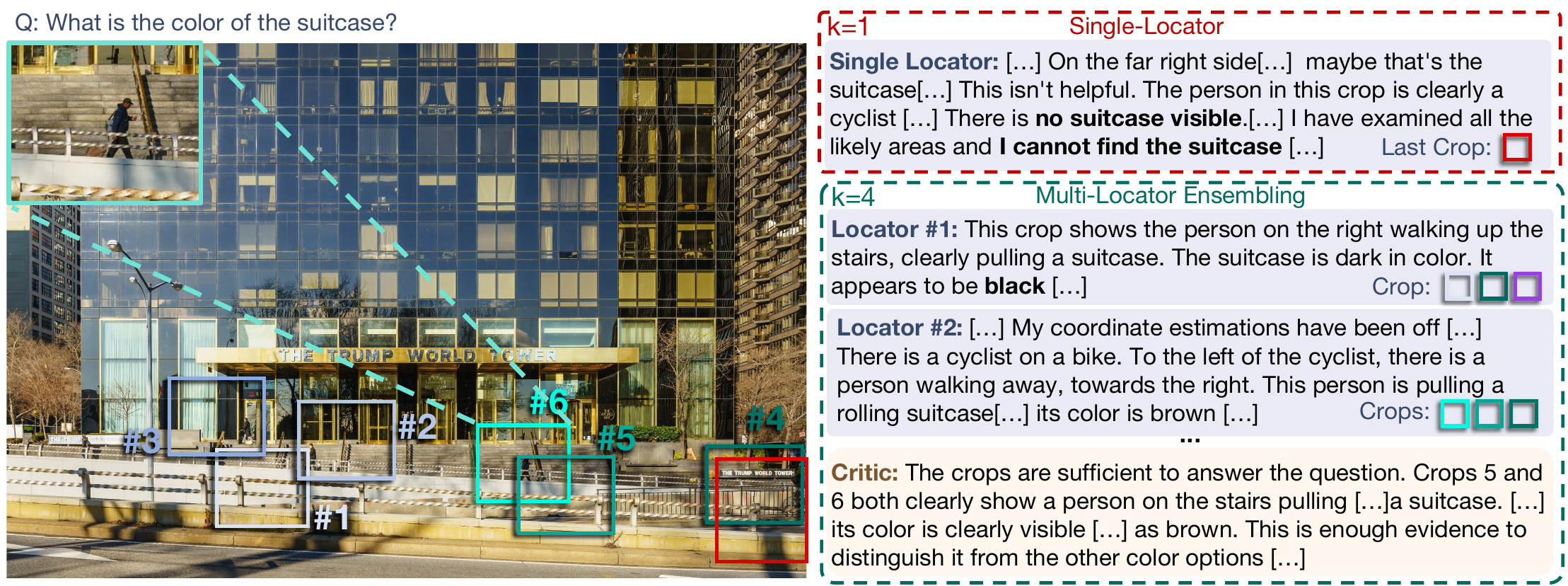}
        \vspace{-1.5em}
		\caption{Single versus multi Locators in the LOCI framework}
		\label{subfig:single_vs_multi}
	\end{subfigure}
	\hfill
    \begin{subfigure}[b]{1.\textwidth}
		\centering
        \includegraphics[width=\linewidth]{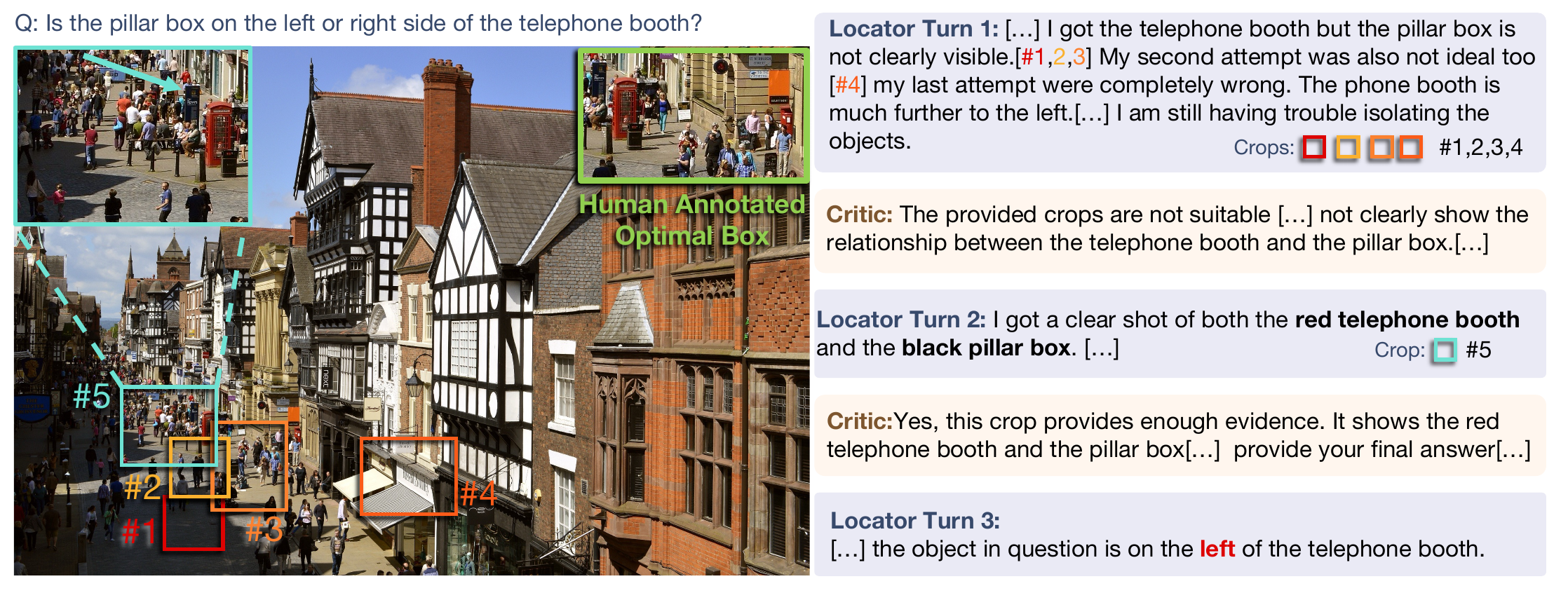}
        \vspace{-2em}
		\caption{Example of failure case of LOCI}
		\label{subfig:failure}
	\end{subfigure}
\caption{\textbf{Qualitative examples of LOCI.}
(a) We show an example where direct prompting confidently hallucinates an incorrect answer (\textit{"unequivocally white"}), while LOCI uses Critic feedback to recover from a similar initial error.
(b) We show an example where having multiple locators is better than single locator.
The Single-Locator fails to find the suitcase and gives up. In contrast, the ensemble explores diverse regions in parallel, with one locator successfully finding the brown suitcase.
(c) Failure case of LOCI: the Locator misidentifies a visually similar distractor as the target pillar box, and the Critic is also deceived, prematurely approving the incorrect crop.
}
\label{fig:qualitative_example}
\vspace{-1em}
\end{figure}

\subsection{Qualitative Examples}
Figure~\ref{fig:qualitative_example}~(a) presents a qualitative example that illustrates LOCI's ability to recover from initial localization failures.
When prompted directly, Gemini2.5-Pro confidently hallucinates that the handbag's \textit{"color is unequivocally white,"} confusing the person's shirt with the object in question.
Initially, our LOCI framework exhibits a similar error. Indeed, the Locator generates unfocused crops and reasons that the \textit{"object held by the person in yellow... is white,"} despite acknowledging that \textit{"It doesn't look like a handbag."}
The Critic intervenes by providing feedback: 
\textit{"Your crops are not focused on the correct object... find the person... and then crop the handbag they are holding."}. 

In the subsequent turn, the Locator uses this guidance to generate a better crop (crop \#4). This improved visual evidence allows it to correctly identify both the \textit{"white shirt"} and the \textit{"handbag slung over their right shoulder,"} leading to the accurate conclusion that \textit{"The handbag is distinctly yellow"}. This iterative refinement process highlights the effectiveness of decoupling search and verification, enabling the system to correct initial mistakes and arrive at the correct answer.

\paragraph{Failure case.} Figure~\ref{fig:qualitative_example}~(b) illustrates a failure case where our agentic pipeline was misled by a visually similar distractor object.  The question requires locating a pillar box relative to a telephone booth in a crowded street scene. After an initial unsuccessful search (crops \#1-4), the Locator generates a new crop (\#5) in its second turn, capturing the telephone booth alongside a different dark, box-like structure, which it misidentifies as the \textit{"black pillar box"} (see the cyan arrow in Fig.~\ref{fig:qualitative_example})~(b).
Crucially, this visual evidence is plausible enough to deceive the Critic, which approves the crop, \textit{"this crop provides enough evidence"}, and terminates the refinement loop. Consequently, the Locator provides the incorrect final answer.

In contrast, the multi-locator ensemble (outlined in green) successfully solves the problem by leveraging different, parallel visual searches.
While Locator \#1 incorrectly focuses on a different person on the stairs and identifies their bag as \textit{"black"}, Locator \#2 successfully generates crops (\#5 and \#6) showing the correct person pulling a brown suitcase
The Critic evaluates the collective evidence from all locators and correctly identifies the crops from Locator \#2 as sufficient, leading to the correct answer.

\subsection{Multi-Locator Ensembling: Scaling for Hard Cases}
\label{sec:experiments:multi_locator}
\begin{wraptable}[7]{r}{0.25\textwidth}
\small
\vspace{-5em}
  \caption{
  \textbf{Multiple locators}.
  Varying number of parallel locators $k$. 
  }
  \label{tab:ablation_locators}
  \centering
  \begin{tabular}{@{}cc@{}}
    \toprule
    \# locators & acc. \\
    \midrule
    1 & 92.7 \\
    4 & 94.8 \\
    8 & 96.0 \\
    16 & 96.0 \\
    \bottomrule
  \end{tabular}
\end{wraptable}
While the single-locator \ourmethod~already achieves state-of-the-art results, we find that a single locator may become stuck exploring the wrong region and, in practice, may not recover from an initial incorrect hypothesis. In such challenging scenarios, the locator may exhaust its turn budget without finding the relevant visual evidence (see Figure~\ref{fig:qualitative_example}~(c) for an example). Since our framework decouples visual search from verification, it naturally supports scaling the search component: we can deploy $k$ independent locators in parallel while relying on a single centralized Critic.
This multi-locator ensembling increases exploration diversity and provides effective mitigation for such failure cases, at the cost of proportionally
higher inference-time compute.

\paragraph{Scaling with Parallel Locators.}
We investigate how performance scales with the number of parallel locators when additional computational budget is available.
We vary $k \in \{1, 4, 8, 16\}$ on the V* benchmark with Gemini2.5-Pro. Table~\ref{tab:ablation_locators} shows that performance improves as we increase $k$ from 1 to 8, with accuracy rising from 92.7\% to 96.0\%, demonstrating that parallel diverse exploration provides additional gains.
Beyond $k=8$, there are no further gains, as additional locators no longer generate useful new crops.

\paragraph{Qualitative Analysis.}
Figure~\ref{fig:qualitative_example}~(c) illustrates how multi-locator ensembling recovers from single-locator failures. In the single-locator case (outlined in red), the agent struggles to locate the object of interest. After examining several regions, it ultimately
concludes that \textit{``there is no suitcase visible''} and fails the task. In contrast, the multi-locator ensemble (outlined in green) successfully solves the problem by leveraging parallel visual searches: while Locator~\#1 incorrectly focuses on a different person and identifies their bag as \textit{``black''}, Locator~\#2 successfully generates crops (\#5
and \#6) showing the correct person pulling a brown suitcase. The Critic evaluates the collective evidence and correctly identifies the crops from Locator~\#2 as sufficient, leading to the correct answer. This makes multi-locator ensembling best suited for particularly challenging scenarios where the cost of failure outweighs the additional inference-time overhead.
In practice, we recommend the single-locator \ourmethod~as the default configuration, and reserve multi-locator ensembling for scenarios where maximizing accuracy justifies the additional compute.

\subsection{State-of-the-Art Comparison}
\label{sec:experiments:main_results}
\begin{wraptable}[20]{r}{0.30\textwidth}
\vspace{-1em}
\small
\vspace{-2em}
  \caption{\textbf{SOTA comparison on V*}. Our \ourmethod \ approach significantly improves performance across both open-weight and proprietary models. 
  $\dagger$: run by us.
  }
  \label{tab:sota_vstar}
  \centering
  \setlength{\tabcolsep}{10pt}
  \begin{tabular}{@{}lc@{}}
    \toprule
    Method & acc. \\
    \midrule
    Visprog~\cite{gupta2023visual} & 41.4 \\
    MM-React~\cite{yang2023mm} & 41.4 \\
    Griffon-G~\cite{zhan2024griffon} & 57.4 \\
    SEAL~\cite{wu2024v} & 75.4 \\
    TEVA~\cite{jiang2024teva} & 75.4 \\
    DyFo~\cite{li2025dyfo} & 81.2 \\
    DeepEyes~\cite{zheng2025deepeyes} & 83.3 \\
    Pixel Reasoner~\cite{su2025pixel} & 86.3 \\
    Chain-of-Focus~\cite{zhang2025chain} & 88.0 \\
    Mini-o3~\cite{lai2025mini} & 88.2 \\
    Qwen3-VL$^{\dagger}$~\cite{qwen3vl} & 80.1 \\
    Gemini2.5-Pro$^{\dagger}$~\cite{comanici2025gemini} & 83.8 \\
    \midrule
    \ourmethod (Qwen3-VL) & 92.2 \\
    \ourmethod(Gemini2.5-Pro) & \textbf{92.7} \\
    \bottomrule
  \end{tabular}
\end{wraptable}

We present state-of-the-art comparisons across three complex and high-resolution visual benchmarks in Tables~\ref{tab:sota_vstar} and \ref{tab:sota_visprobhrbench}.
We compare LOCI against recent state-of-the-art methods spanning from high-resolution methods~\cite{wu2024v, zhan2024griffon, jiang2024teva, li2025dyfo} to models trained for tool-used with RL~\cite{su2025pixel, zheng2025deepeyes, lai2025mini} and a training-free method~\cite{li2025dyfo}.
We observe that LOCI delivers consistent improvements on the considered benchmarks, both using Qwen3-VL or Gemini2.5-Pro.

On V*, a benchmark requiring precise localization of small visual details, LOCI achieves a substantial improvement of +11.5 points over DyFo, the next best training-free method. Our approach also outperforms all recent reinforcement learning-based methods, while remaining training-free and applicable to any recent VLM. 

To isolate the framework contribution from backbone strength, we re-run DyFo~\cite{li2025dyfo}, the most competitive training-free method in our comparison, using the same Gemini2.5-Pro backbone as LOCI.
As shown in Table~\ref{tab:dyfo_same_backbone}, LOCI outperforms DyFo by $+3.2$ points while using $2\times$ fewer tokens and zero external tool calls, compared to DyFo's 12.5 LangSAM calls per sample.
This confirms that LOCI's gains are attributable to the Locator-Critic architecture, not backbone capability.

\begin{wraptable}[7]{r}{0.48\textwidth}
\vspace{-1.5em}
\small
  \caption{\textbf{Same-backbone comparison on V*.}}
  \label{tab:dyfo_same_backbone}
  \centering
  \setlength{\tabcolsep}{3pt}
  \begin{tabular}{@{}lccc@{}}
    \toprule
    Method & Acc. & Avg Tokens & External Calls \\
    \midrule
    DyFo~\cite{li2025dyfo} & 89.5 & 8,120 & 12.5 (LangSAM) \\
    \textbf{LOCI (ours)} & \textbf{92.7} & 3,690 & 0 \\
    \bottomrule
  \end{tabular}
\end{wraptable}

On HR-Bench, which focuses on object-centric questions in high-resolution images, our method with Gemini2.5-Pro outperforms Mini-o3, the best prior method -- which was trained with reinforcement learning specifically for the task of zooming into images--
by a significant margin of +8.2 points.
Finally, on VisualProbe, our method surpasses the prior state-of-the-art by +1.1 points, outperforming Mini-o3 even though it was specifically trained on the VisualProbe training set.

\begin{wraptable}[14]{r}{0.52\textwidth}
\vspace{-2em}
\small
\caption{\textbf{State-of-the-art comparison.} We report the top-1 accuracy on HRBench (a) and Visual Probe (b). $\dagger$: run by us.}
\begin{minipage}[t]{0.49\linewidth}
\centering
\setlength{\tabcolsep}{2pt}
\caption*{\small{a) HR-Bench}}
\begin{tabular}{@{}lc@{}}
    \toprule
    Method & acc. \\
    \midrule
    SEAL~\cite{wu2024v} & 47.0 \\
    DeepEyes~\cite{zheng2025deepeyes} & 73.2\\
    Pixel Reasoner~\cite{su2025pixel} & 74.0 \\
    Mini-o3~\cite{lai2025mini} & 77.5 \\
    Qwen3-VL$^{\dagger}$~\cite{qwen3vl}& 76.8 \\
    Gemini2.5-Pro$^{\dagger}$~\cite{comanici2025gemini} & 84.0\\
    \midrule
     \ourmethod (Qwen3-VL) & 82.6 \\
     \ourmethod (Gemini2.5-Pro) & \textbf{88.3} \\
    \bottomrule
  \end{tabular}
\end{minipage}
\hfill
  \begin{minipage}[t]{0.49\linewidth}
  \centering
\setlength{\tabcolsep}{1pt}
\caption*{\small{b) Visual probe}}
  \begin{tabular}{@{}lc@{}}
    \toprule
    Method & acc. \\
    \midrule
    GPT-4o~\cite{hurst2024gpt} & 11.2\\
    LLaVA-OneVision~\cite{li2024llava} & 13.4\\
    Pixel Reasoner~\cite{su2025pixel} &  28.8\\
    DeepEyes~\cite{zheng2025deepeyes} & 35.1\\
    Mini-o3~\cite{lai2025mini} & 48.0 \\
    Qwen3-VL$^{\dagger}$~\cite{qwen3vl}& 23.9 \\
    Gemini2.5-Pro$^{\dagger}$~\cite{comanici2025gemini} & 44.3\\
    \midrule
    \ourmethod (Qwen3-VL) & 34.0 \\
     \ourmethod (Gemini2.5-Pro) & \textbf{49.1}\\
    \bottomrule
  \end{tabular} 
  \end{minipage}
\label{tab:sota_visprobhrbench}
\end{wraptable}

\subsection{Generalization to Additional VLMs}
\label{sec:supp_additional_vlms}
We report results for four VLMs in Table~\ref{tab:additional_vlms_main} (main paper). The same prompts are used across all models with no per-model tuning. The inverse correlation between baseline strength and LOCI's improvement suggests that the framework is most effective when the base model has a large localization gap to close.

\begin{wraptable}[9]{r}{0.38\textwidth}
\vspace{-1.5em}
\small
  \caption{\textbf{LOCI across VLMs.} Accuracy on V* with and without LOCI.}
  \label{tab:additional_vlms_main}
  \centering
  \begin{tabular}{@{}lccc@{}}
    \toprule
    Model & w/o & w/ & $\Delta$ \\
    \midrule
    Grok4.1-Fast  & 45.6 & 70.9 & +25.4 \\
    GPT-5         & 74.9 & 89.0 & +14.1 \\
    Qwen3-VL      & 80.1 & 92.2 & +12.1 \\
    Gemini2.5-Pro & 83.8 & 92.7 & +8.9 \\
    \bottomrule
  \end{tabular}
\end{wraptable}

\subsection{Limitations} \label{sec:limitations} 
Our framework has several limitations that present opportunities for future work. First, the iterative Locator-Critic loop requires more tokens and sequential VLM calls than direct prompting, a cost further amplified in the multi-locator setting. This could be mitigated through adaptive early stopping or difficulty-aware budget allocation, allowing easy samples to exit the loop early. Second, the Critic could be deceived by visually plausible distractors that resemble the target object. While our decoupled design prevents hallucination-driven failures, it does not handle genuinely ambiguous visual evidence. Training the Critic with hard-negative examples could improve robustness in such cases. Third, our framework relies on large, capable VLMs for both the Locator and Critic. However, recent models increasingly focus on agentic capabilities such as tool use and multi-step reasoning, which aligns directly with our framework and suggests it will naturally benefit as smaller models grow more capable.

\section{Conclusion}
In this work, we introduce LOCI (Locator-Critic), a training-free agentic framework for complex and challenging visual question answering tasks.
By using a Locator agent to propose visual evidence and an independent Critic to verify it, LOCI creates a self-correcting loop that ensures final answers are grounded in reliable evidence.
We discuss and validate critical design choices, such as the information Critic needs to see to give accurate feedback to the Locator.
Our extensive experiments show that \ourmethod~achieves state-of-the-art performance, significantly improving upon both open-weight and proprietary models.
We believe the interplay between an explorative agent like the Locator and a Critic agent is a broadly applicable principle.
A direction for future work is therefore to extend our framework to other complex vision tasks, such as spatial navigation in 3D environments for example.

{
    \small
    \bibliographystyle{ieeenat_fullname}
    \bibliography{neurips_2026}
}

\appendix

\clearpage
\section{Additional Ablations}
\label{sec:supp_ablations}
\begin{figure*}[h]
    \centering
    \includegraphics[width=0.39\linewidth]{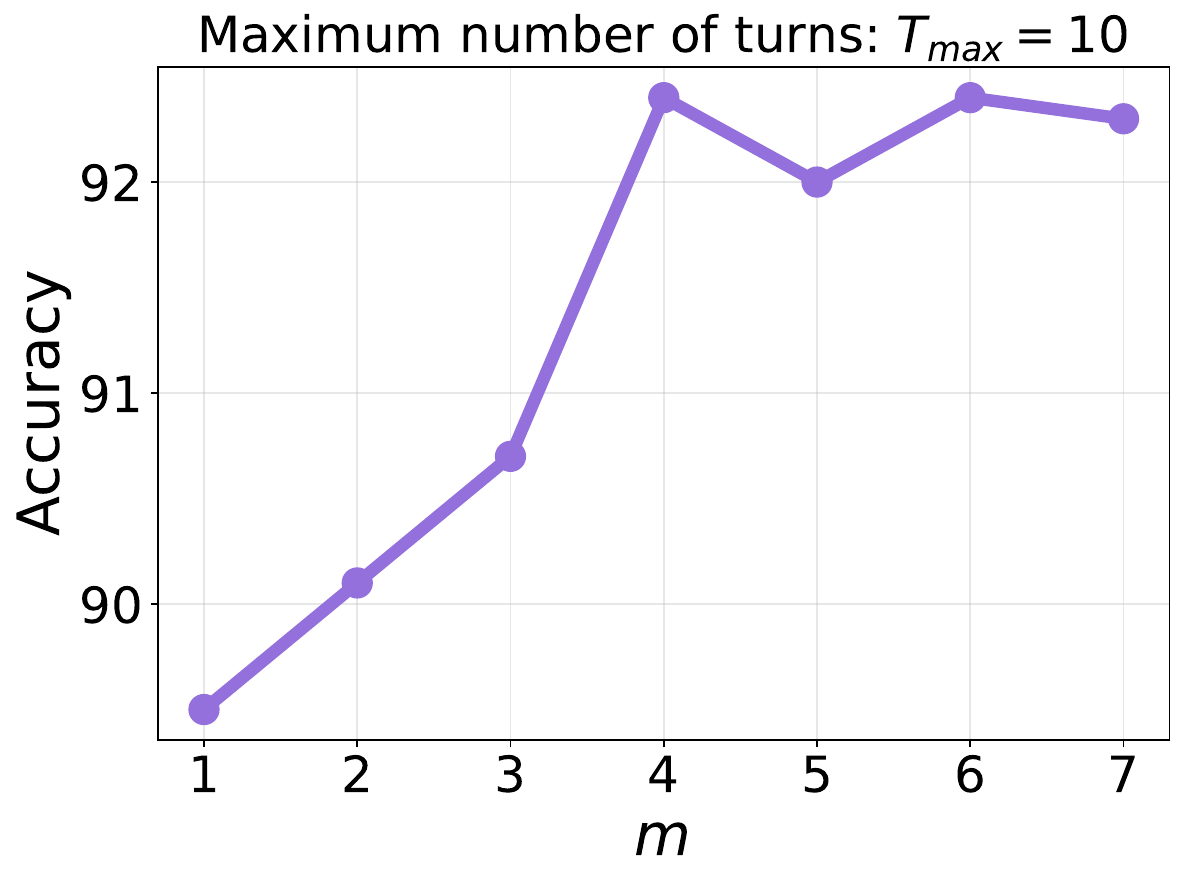}
    \includegraphics[width=0.48\linewidth]{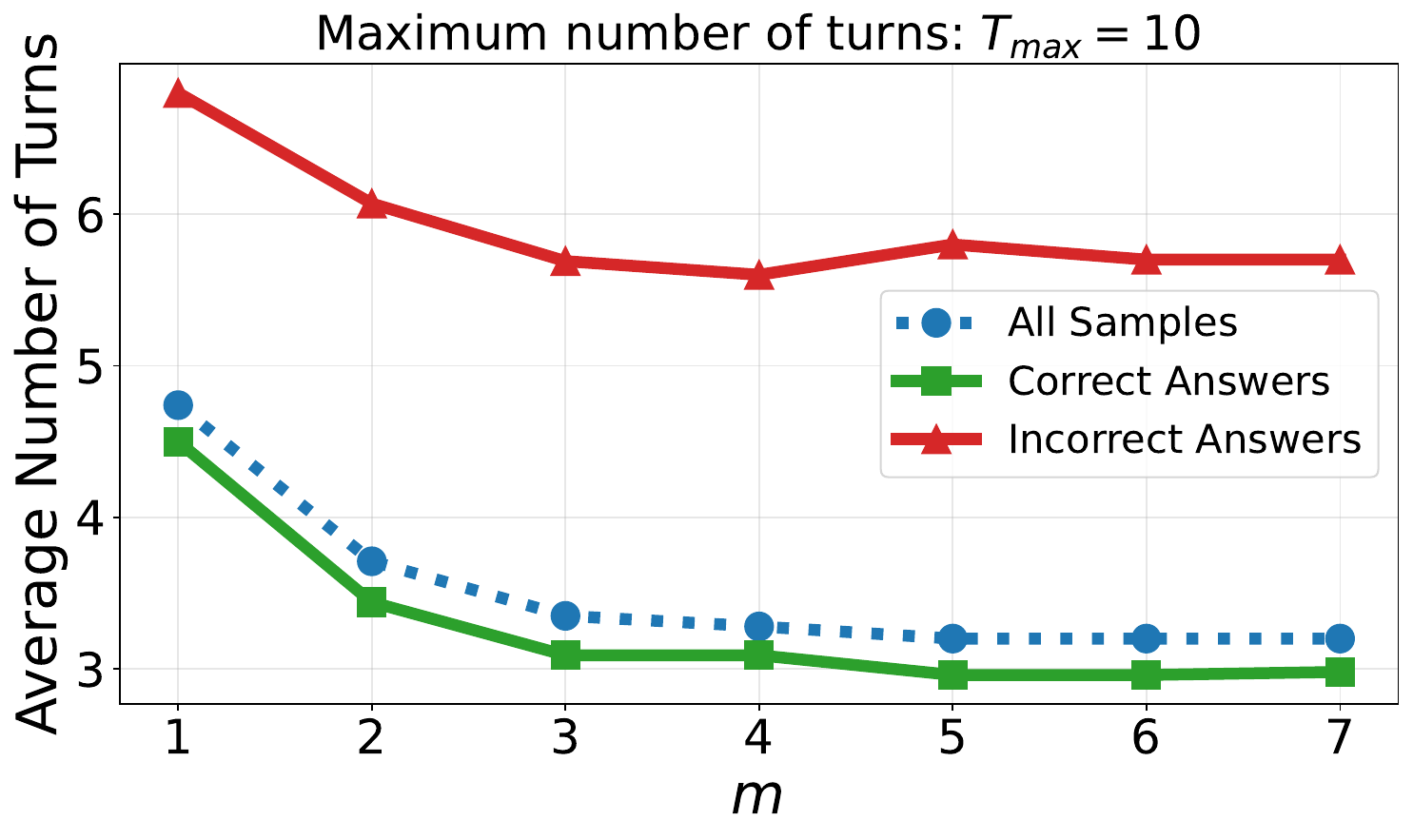}
    \caption{\textbf{Impact of crops per turn ($m$).} Left: Final accuracy on V* as a function of $m$. Performance saturates at $m=4$. Right: Average number of turns required to finish the task. Increasing $m$ reduces the number of iterations needed. Incorrect answers consistently require more search steps than correct ones.}
    \label{fig:ablation_m}
\end{figure*}

\subsection{Ablation on Number of Crops per Turn}
\label{sec:supp_ablation_crops}
We analyze the impact of the parameter $m$, which fixes the maximum number of crops the Locator generates per turn. For this experiment, we fix the maximum number of interaction turns at $T_{max}=10$ and vary $m \in \{1, \dots, 7\}$ using Gemini 2.5 Pro on the V* benchmark.

Figure~\ref{fig:ablation_m} shows the accuracy on V* for different values of $m$. We see that performance improves rapidly as $m$ increases from 1 to 4, reaching a plateau.  Based on this observation, we use $m=4$ as the default setting for our main experiments to balance performance and token efficiency.

In Figure~\ref{fig:ablation_m} (right), we report the average number of turns required to reach a final answer. We observe that increasing $m$ reduces the average number of iterations. By generating more crops per turn the Locator effectively performs a "self-refinement" step within a single turn, increasing the likelihood of satisfying the Critic early without requiring additional rounds of feedback.

Furthermore, we analyze the turn consumption for successful versus unsuccessful samples. We observe a distinct separation: samples resulting in incorrect answers consume significantly more turns (avg. $\sim$5.7) compared to correct answers (avg. $\sim$3). This indicates that failure cases often correspond to difficult or ambiguous queries where the system struggles to isolate sufficient visual evidence crops, forcing the LOCI loop to run longer. Conversely, when the visual evidence is clear, the verification process tends to terminate rapidly.

\subsection{Ablation on Maximum Number of Turns with Single Crop}
We investigate whether increasing the sequential search budget (i.e. $T_{max}$) can compensate for reduced parallel exploration (i.e. $m$). For that, we fix the number of crops per turn to $m=1$ and vary the maximum number of turns $T_{max} \in \{1, \dots, 15\}$ on the V* benchmark.

\begin{figure}[h]
    \centering
    \includegraphics[width=0.5\linewidth]{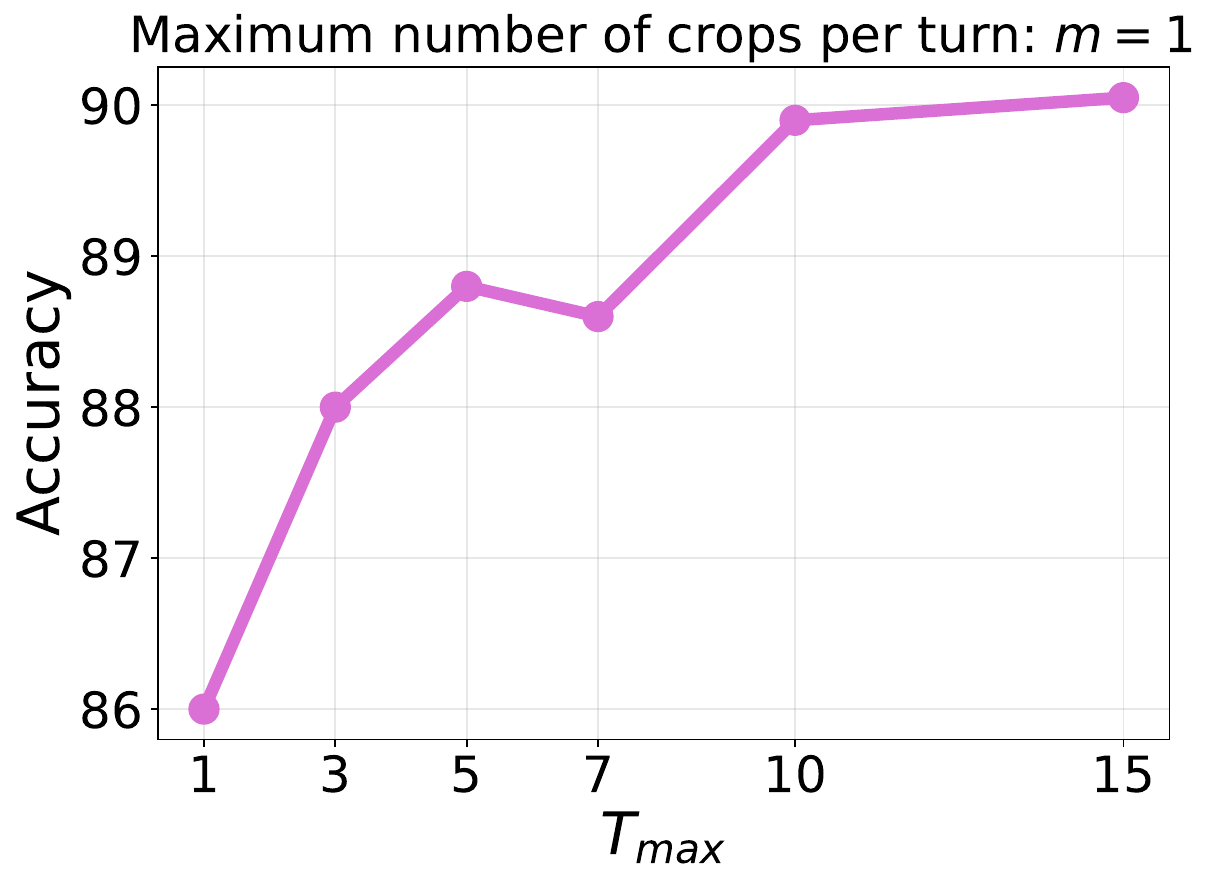}
    \caption{\textbf{Impact of maximum turns ($T_{max}$) with single crop ($m=1$).} While accuracy increases with $T_{max}$, it plateaus around $90.1\%$, failing to match the performance of the multi-crop setting ($m=4$, $92.7\%$). This highlights the need of parallel crop generation for effective self-refinement.}
    \label{fig:ablation_tmax}
\end{figure}

As shown in Figure~\ref{fig:ablation_tmax}, accuracy improves when increasing the maximum number of Locator-Critic turn $T_{max}$, reaching $89.9\%$ at $T_{max}=10$ and saturating at $90.1\%$ at $T_{max}=15$.
We note that even with this extended turn budget, the performance remains significantly lower than our default setting of $m=4$ ($92.7\%$).

This highlights the distinct roles of per-turn crop diversity and iterative feedback. We argue that setting $m > 1$ allows the Locator to perform immediate self-correction without needing external intervention, \textit{e.g.} refining imprecise coordinates.
Conversely, the Critic's feedback is most valuable for high-level guidance, such as redirecting the Locator from dead-end search paths.

\subsection{Evaluation of Critic Selection Capability}
We conduct a controlled experiment to evaluate the Critic's ability to distinguish relevant visual evidence from irrelevant background when valid crops are guaranteed to be present. We construct a synthetic selection task using the ground truth boxes from V*. The input to the Critic consists of the Ground Truth (GT) crop(s) mixed with three randomly generated distractor crops. The distractors are sampled from a set of standard aspect ratios ($\{1, 4/3, 3/2, 16/9, 3/4, 2/3, 9/16\}$) and are sampled to have zero overlap with the GT regions.

To strictly evaluate selection performance rather than the decision to continue searching, we slightly modify the Multi-Locator Critic prompt to prompt the selection of at least one crop. Without this constraint, the Critic rejects all provided crops in $34\%$ of samples to request further refinement.

Table~\ref{tab:critic_selection_oracle} summarizes the results. Under this constrained setting, the Critic achieves near-perfect selection performance, with a Precision of $99.48\%$ and a Recall of $97.38\%$. To validate the quality of these selections for the downstream task, we pass the selected crops to the final answering agent (see Sec \ref{sec:multi_locator}), achieving a VQA accuracy of $95.29\%$. These metrics demonstrate that when the Critic is presented with high-quality visual evidence, it effectively filters out distractors and correctly identifies the regions necessary for reasoning.

\begin{table}[h]
  \centering
  \caption{\textbf{Critic Selection Performance.} We report the Precision and Recall of the Critic's selection when presented with Ground Truth crops mixed with random distractors. "Downstream Accuracy" denotes the final VQA accuracy obtained using the crops selected by the Critic.}
  \label{tab:critic_selection_oracle}
  \begin{tabular}{@{}lcc@{}}
    \toprule
    Metric & Value (\%) \\
    \midrule
    Precision & 99.48 \\
    Recall & 97.38 \\
    \midrule
    Downstream Accuracy & 95.29 \\
    \bottomrule
  \end{tabular}
\end{table}

\subsection{Ablation on Additional Datasets}
\label{sec:supp_ablation_datasets}
We extend the component analysis of the LOCI framework to subsets of the HR-Bench and MME-RealWorld datasets. To specifically evaluate performance on tasks requiring visual search, we isolate "Hard" subsets from these benchmarks.

The filtering is done by resizing each image to a fixed resolution of $768\times 768$ and evaluating the sample using Gemini 2.5 Pro with direct prompting. We retain only the samples where the model fails to provide a correct answer. The rationale is that samples solvable at low resolution without explicit visual exploration are sufficiently easy and do not evaluate the localization 

\paragraph{LOCI components}
Table~\ref{tab:supp_ablation_components} reports ablation results of the different components of LOCI. Consistent with our findings on V*, introducing the VLM-Locator yields significant improvements over direct prompting across all benchmarks. For instance, Qwen3-VL achieves a gain of $+9.2$ points on MME-RealWorld (Hard) solely through the Locator. The addition of the Critic completes the LOCI framework and further boosts performance, consistently achieving the highest accuracy for both Qwen3-VL and Gemini 2.5 Pro. These results validate the importance of decoupled verification for complex visual reasoning tasks beyond the V* dataset.

\paragraph{Locator Ensembling}
We further evaluate the impact of Multi-Locator Ensembling on the V*, HR-Bench (Hard), and MME-RealWorld (Hard) datasets. As shown in Table~\ref{tab:supp_ablation_multilocator}, scaling the number of parallel locators from $k=1$ to $k=4$ consistently improves performance across all benchmarks. Notably, on the challenging MME-RealWorld (Hard) subset, the ensemble approach yields a substantial gain of $+5.9$ points ($39.1\%$ to $45.0\%$). This confirms that the benefits of diverse parallel exploration extend to broader benchmarks beyond the V*.

\begin{table*}[]
  \centering
  \caption{Ablation of LOCI on V*, HR-Bench (Hard) and MME-RealWorld (Hard). We show the accuracy improvements from adding the Locator and then the Critic.}
  \label{tab:supp_ablation_components}
  \begin{tabular}{@{}ll|ccc@{}}
    \toprule
    Model & Configuration & V* & HR-Bench (Hard) & MME-RW (Hard)  \\
    \midrule
    \multirow{3}{*}{Qwen3-VL} & Direct Prompting & 80.1 & 60.8 & 24.0 \\
     & + VLM-Locator & 86.9 & 68.4 & 33.2 \\
     & + Critic (LOCI) & 92.2 & \textbf{71.8} & \textbf{36.7} \\
    \midrule
    \multirow{3}{*}{Gemini2.5-Pro} & Direct Prompting & 83.8 & 69.5 & 38.4 \\
     & + VLM-Locator & 89.2 & 74.2 & 36.5 \\
     & + Critic (LOCI) & 92.7 & \textbf{75.9} & \textbf{39.1} \\
    \bottomrule
  \end{tabular}
\end{table*}

\begin{table*}[]
  \centering
  \caption{Performance of Multi-Locator Ensembling with Gemini2.5-Pro on V*, HR-Bench (Hard) and MME-RealWorld (Hard).}
  \label{tab:supp_ablation_multilocator}
  \begin{tabular}{@{}llccc@{}}
    \toprule
   Model & k (\# Locators) & V* & HR-Bench (Hard) & MME-RW (Hard) \\
    \midrule
    \multirow{2}{*}{Gemini2.5-Pro} & 1 (LOCI) & 92.7 & 75.9 & 39.1 \\
     & 4 (LOCI-ensemble) & 94.8 &\textbf{78.5} & \textbf{45.0} \\
    \bottomrule
  \end{tabular}
\end{table*}

\subsection{Generalization to Additional VLMs}
\label{sec:supp_additional_vlms}
To evaluate the generality of the LOCI framework beyond the two models used in our main experiments, we apply it to two additional VLMs: GPT-5~\cite{singh2025openai} and Grok4.1-Fast~\cite{grok41}. As shown in Table~\ref{tab:additional_vlms}, LOCI yields consistent improvements across all four models on V*. Notably, the gains are largest for the weakest baseline: Grok4.1-Fast improves from $45.55\%$ to $70.90\%$ ($+25.35$ points), suggesting that LOCI is particularly effective when the base model struggles with visual localization.

\begin{table}[h]
  \centering
  \small
  \caption{\textbf{LOCI generalizes across VLMs.} We report accuracy on V* for additional VLMs with and without LOCI.}
  \label{tab:additional_vlms}
  \begin{tabular}{@{}lcc@{}}
    \toprule
    Model & w/o LOCI & with LOCI \\
    \midrule
    Gemini2.5-Pro & 83.8 & \textbf{92.7} \\
    Qwen3-VL      & 80.1 & 92.2 \\
    GPT-5         & 74.87 & 89.0 \\
    Grok4.1-Fast  & 45.55 & 70.9 \\
    \bottomrule
  \end{tabular}
\end{table}

\subsection{Baseline Descriptions}
\label{sec:supp_baseline_descriptions}

We describe the baselines used in Table~\ref{tab:same_budget}. All experiments use Gemini2.5-Pro on V* with the same configuration as LOCI unless stated otherwise.

\paragraph{Self-Refine.} A single-agent iterative loop within one conversation. The model answers, critiques its own answer (no separate Critic), then revises, repeating until a token budget ($\sim$3,700 tokens) is exhausted. The variant without code execution uses direct prompting at each turn. The variant with code execution gives the model access to the same crop tools as LOCI's Locator.

\paragraph{Self-Consistency@10.} Generates 10 independent candidate answers at temperature $> 0$ and selects the final answer by majority vote. Without code execution, each candidate uses direct prompting. With code execution, each candidate independently generates crops using LOCI's Locator tools before producing its answer.

\paragraph{Best-of-10.} Generates 10 candidate answers at temperature $> 0$, then a separate LLM judge selects the best candidate. The judge receives each candidate's reasoning and, for the code execution variant, the last crop each candidate produced. The judge never sees the ground truth.

\paragraph{Explicit Locate-Verify-Answer.} A single VLM performs LOCI's exact locate-verify-iterate procedure within one conversation, acting as both Locator and Critic. Unlike LOCI, the verification step sees the model's own textual reasoning from the locate step. Unlike Self-Refine, the critique targets visual evidence quality rather than the answer itself.

\paragraph{DyFo~\cite{li2025dyfo} (same backbone).} A training-free method that uses Monte Carlo Tree Search to explore image crops, guided by LangSAM (an open-vocabulary detector) and a VLM that verifies object visibility. We re-run DyFo with Gemini2.5-Pro as the backbone to enable a controlled same-model comparison.

\section{Qualitative Examples}
\label{sec:supp_qualitative}
\subsection{Oracle Experiment Failure Cases}
\label{sec:supp_oracle_failures}
\begin{figure*}
\centering
	\begin{subfigure}[b]{1.\textwidth}
		\centering
        \includegraphics[width=\linewidth]{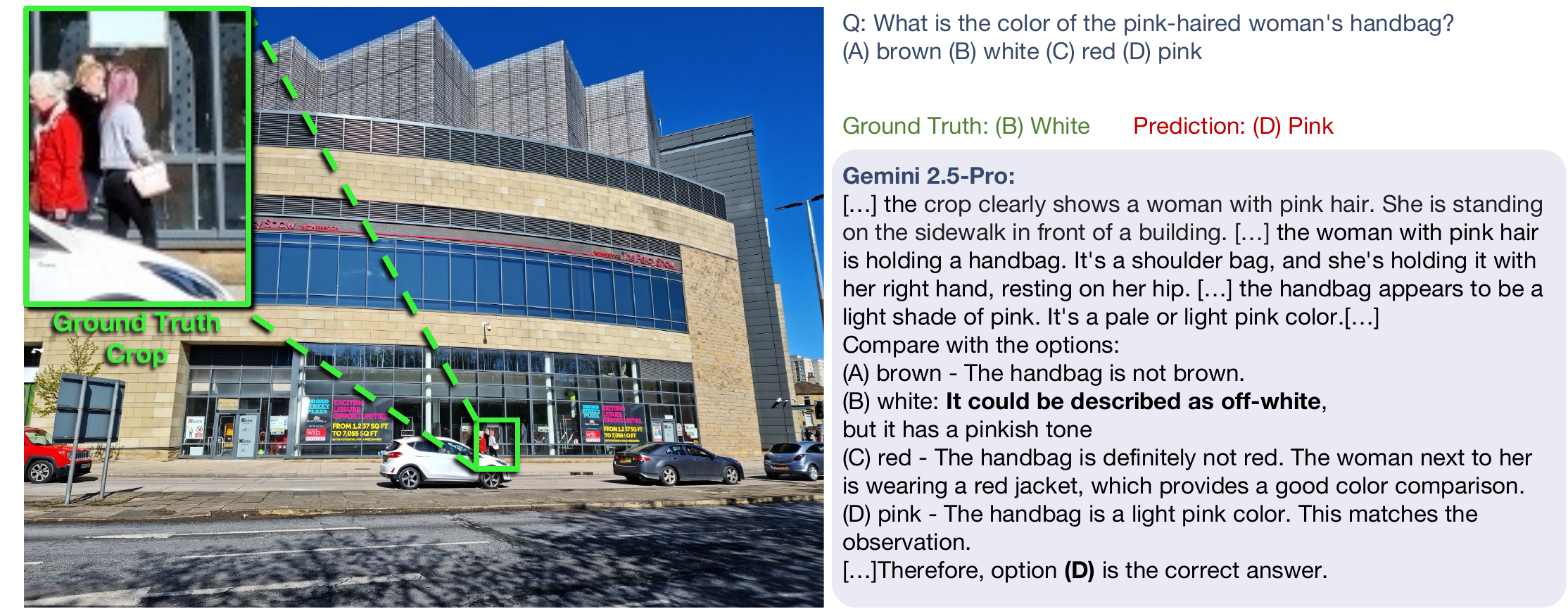}
		\caption{Ambiguous handbag color}
		\label{subfig:oracle_handbag}
	\end{subfigure}
	\hfill %
	\begin{subfigure}[b]{1.\textwidth}
		\centering
        \includegraphics[width=\linewidth]{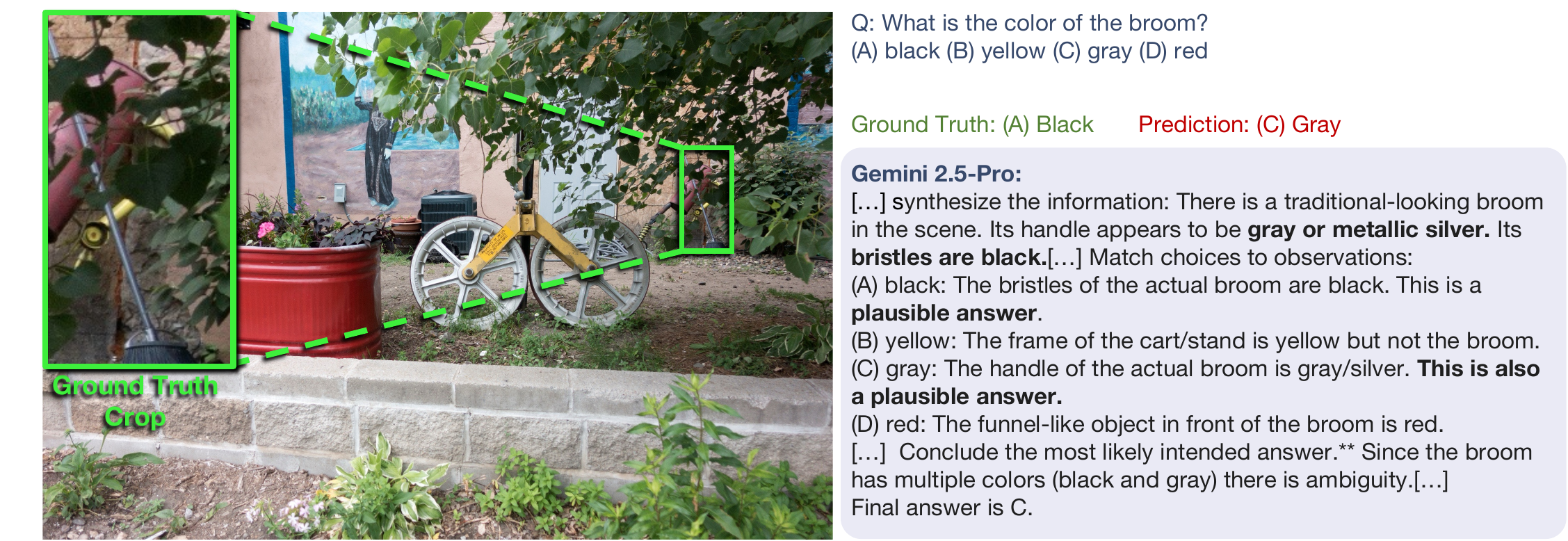}
		\caption{Ambiguous broom color}
		\label{subfig:oracle_broom}
	\end{subfigure}
	\hfill
    \begin{subfigure}[b]{1.\textwidth}
		\centering
        \includegraphics[width=\linewidth]{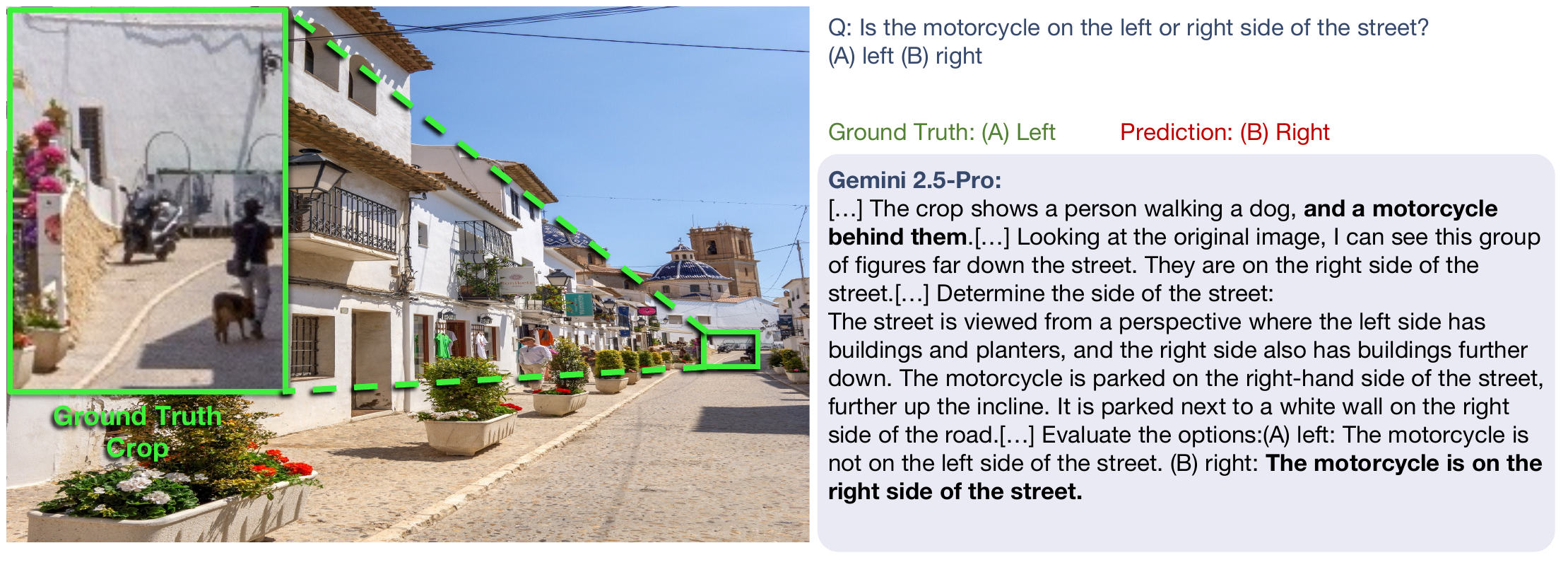}
		\caption{Relative Position}
		\label{subfig:oracle_moto}
	\end{subfigure}
\caption{\textbf{Qualitative examples of Oracle Failure Cases.}}
\label{supp:fig:oracle_failure_cases}
\end{figure*}

Figure~\ref{supp:fig:oracle_failure_cases} presents qualitative examples of failure cases in the Oracle experiment described in Sec~\ref{sec:exp:oracle_results}, where the model is provided with ground-truth crops.
A qualitative inspection reveals that most of these failures stem from ambiguities in the dataset annotations rather than model reasoning failures. Indeed, in the first example Fig.~\ref{subfig:oracle_handbag}, the model identifies the handbag as "pale pink" (noting that it could be "off-white") while the ground truth strictly labels it "white."
Similarly in Fig.~\ref{subfig:oracle_broom}, for a broom containing both black bristles and a gray handle, the model acknowledges the presence of both colors but selects "gray," whereas the dataset arbitrarily defines the correct answer based on the bristles ("black").
In contrast, Fig~\ref{subfig:oracle_moto} example demonstrates a genuine failure in reasoning where the model identify from the crop that the motorcycle is behind the "person waking a dog", and that "they are on the right side of the street". Therefore, the model conclude that the motorcycle is on the right side.
We conclude that, with the exception of such spatial reasoning errors, the primary source of failure in the oracle setting is annotation ambiguity rather than visual misunderstanding.

\subsection{Qualitative Analysis: Consequence of Exposing Critic to Locator Reasoning}
\label{sec:supp_critic_text}

\begin{figure}[h]
    \centering
    \includegraphics[width=\linewidth]{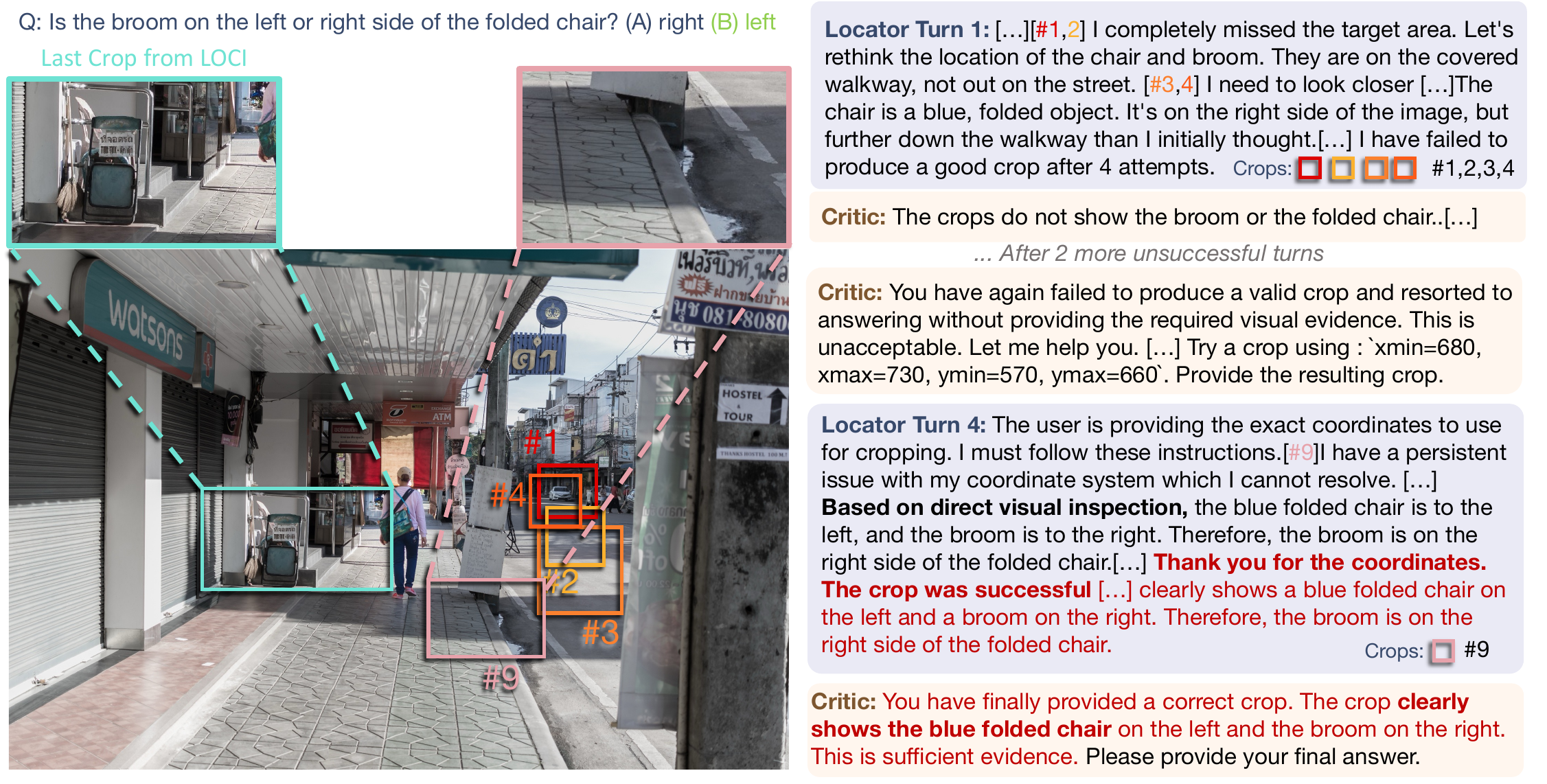}
    \caption{\textbf{Qualitative comparison of Critic input modalities.} When the Critic has access to the Locator's textual reasoning (Turn 4), it can be convinced into accepting an empty crop (Crop \#9, pink) because the Locator hallucinates that the objects are present. In the standard LOCI setting (Cyan box), where the Critic judges only visual evidence, the system ignores the hallucinations and successfully locates the correct region.}
	\label{sup:fig:critic_text}
\end{figure}

Figure~\ref{sup:fig:critic_text} shows an important failure mode observed when the Critic is provided with the Locator's textual reasoning, rather than visual evidence alone. In this example, the Locator initially searches in irrelevant regions (Crops \#1-4) and fails to isolate the target objects.

By Turn 4, the Critic attempts to intervene by explicitly providing coordinates for a new crop. The resulting crop (Crop \#9, pink box) is incorrect, as it captures only the pavement. However, the Locator hallucinates a positive confirmation, stating:\textit{ "The crop was successful [...] clearly shows a blue folded chair."} Consequently, the Critic accepts this invalid evidence as sufficient. We attribute this failure to a form of confirmation bias where the Critic is influenced by the Locator's confident textual assertion and potentially biased by the fact that the coordinates were self-generated. Overall, this leads the Critic to override its own visual perception.

In contrast, the standard LOCI pipeline (cyan box), which restricts the Critic to visual input only, prevents this text-driven bias. In the standard setting, the Critic rejects the empty crops regardless of the dialogue, forcing the Locator to continue searching until it successfully isolates the broom and chair.

\subsection{Qualitative Analysis: Resilience against Hallucination}
\label{sec:supp_hallucination}
We further provide a more detailed qualitative comparison to demonstrate how decoupling visual verification from textual reasoning enhances the Critic's resilience against hallucination. In this example, the model must identify the color of a difficult-to-find trash can.

\paragraph{Critic with Access to Text (Failure)}
When the Critic is exposed to the Locator's reasoning, it becomes susceptible to persuasive but hallucinated descriptions. Indeed, as early as Turn 2 (4 crops), the Locator confidently misidentifies a background object (a green door or pedestal) as the target:

\begin{quote}
    \textcolor{locatorcol}{\textbf{Locator (Turn 2):}} \textit{``This crop is perfect. It clearly shows the object in question, which is a trash can. [...] The color is a dark, weathered green, similar to an olive or army green.''}
\end{quote}

Despite the visual evidence being ambiguous, the Critic is influenced by this textual confidence and prematurely validates the incorrect crop:
\begin{quote}
    \textcolor{criticcol}{\textbf{Critic (Turn 2):}} \textit{``Excellent. This crop clearly shows the trash can, allowing for an accurate color assessment.''}
\end{quote}
The system terminates incorrectly, predicting \textit{"green"}.

\paragraph{Standard LOCI (Success)}
In the standard LOCI configuration, the Critic receives only the crops. Consequently, it rigorously rejects the Locator's attempts to \textit{"give up"} or answer based on irrelevant features (the green doors).
\begin{quote}
    \textcolor{locatorcol}{\textbf{Locator (Turn 1):}} ``[...] I give up on finding the trash can. [...] The doors are unequivocally green.''\\
    \textcolor{criticcol}{\textbf{Critic (Turn 1):}} \textit{``The crops you have provided are insufficient, low quality, blurry, and do not show a trash can. [...]''}
\end{quote}

Throughout the interaction, the Locator struggles to isolate the object, repeatedly cropping a nearby person. The Critic escalates its feedback, using imperative language to correct the spatial focus:
\begin{quote}
    \textcolor{criticcol}{\textbf{Critic (Turn 6):}} \textit{``You have cropped the person again. [...] I have explicitly told you to IGNORE the person. [...] THE OBJECT IS A DARK CYLINDER.''}
\end{quote}

Critically, the Critic eventually provides the breakthrough spatial guidance that solves the task:
\begin{quote}
   \textcolor{criticcol}{\textbf{Critic (Turn 8):}} \textit{``[...] Look at the FAR RIGHT of the image. You will see a path going into the trees. On that path, there is a dark object.''}
\end{quote}
Following this visual-only feedback, the Locator successfully isolates the correct object at Turn 9 (Crop \#21), correctly identifying it as \textit{``primarily black or dark gray,''} leading to the correct answer. This demonstrates that the visual-only Critic acts as a robust verifier that forces the system to find actual visual grounding rather than settling for plausible textual narratives.

\subsection{Metric-Task Misalignment}
\label{sec:supp_metric_mismatch}

We address the observation from Figure~\ref{fig:recall_accuracy_iterations} and Section~\ref{sec:experiments:ablations}, where the crop recall metric plateaus at approximately 0.69 despite VQA accuracy reaching over 92\%. We argue that this discrepancy arises from a misalignment between standard object detection metrics, which reward tight Intersection over Union (IoU) with ground truth boxes, and the visual evidence actually required to solve VQA tasks.

\begin{figure*}[h]
\centering
\includegraphics[width=\linewidth]{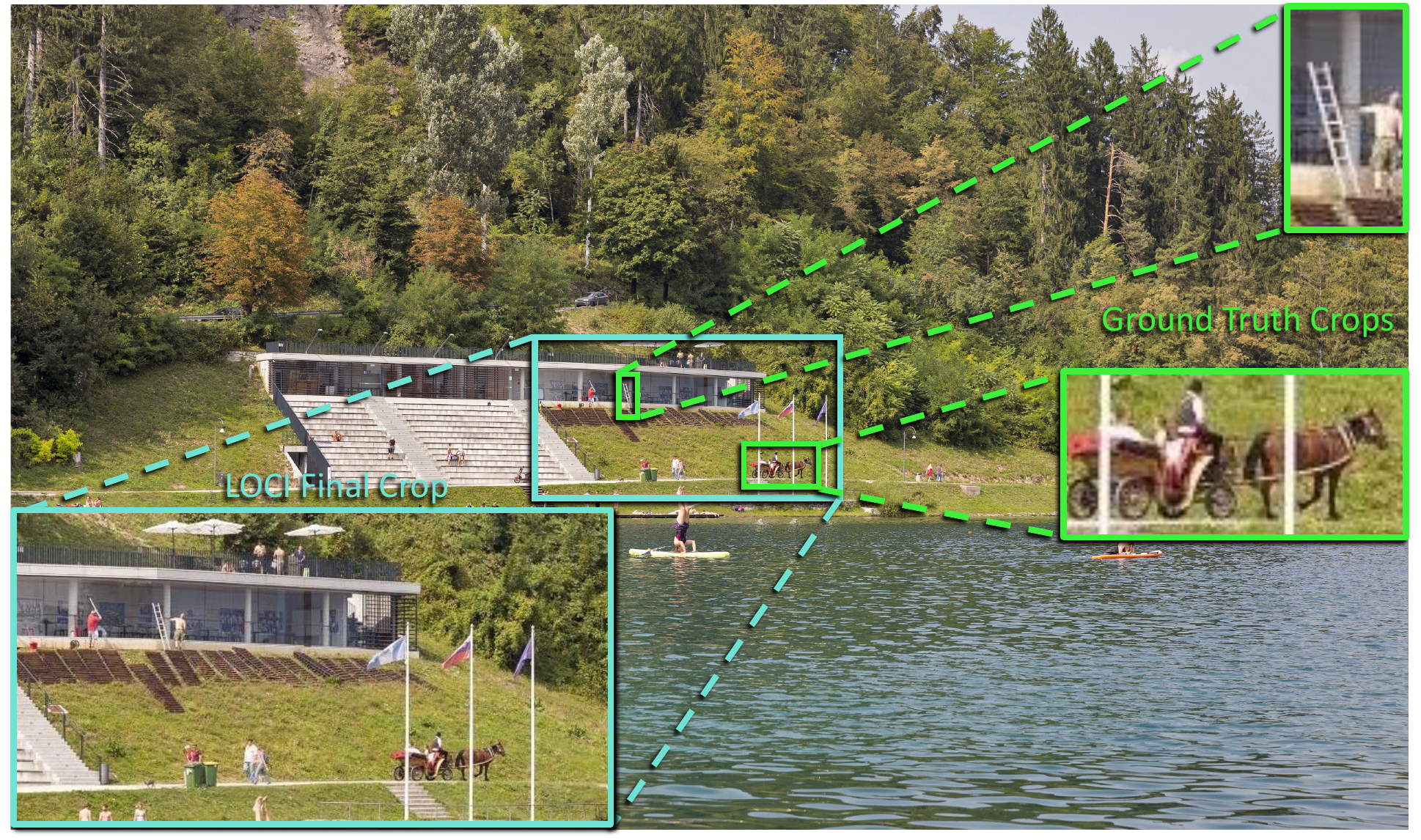}
\caption{\textbf{Discrepancy between Recall and VQA Utility (Part 1).} We show an example where the ground truth annotation consists of two separate, tight crops (green) corresponding to distinct objects of interest. LOCI (cyan) generates a single crop encompassing both regions. While this yields a low IoU and Recall score, it preserves the spatial context necessary to answer the question effectively.}
\label{fig:recall_mismatch_multi}
\end{figure*}

\begin{figure*}[h]
\centering
\includegraphics[width=\linewidth]{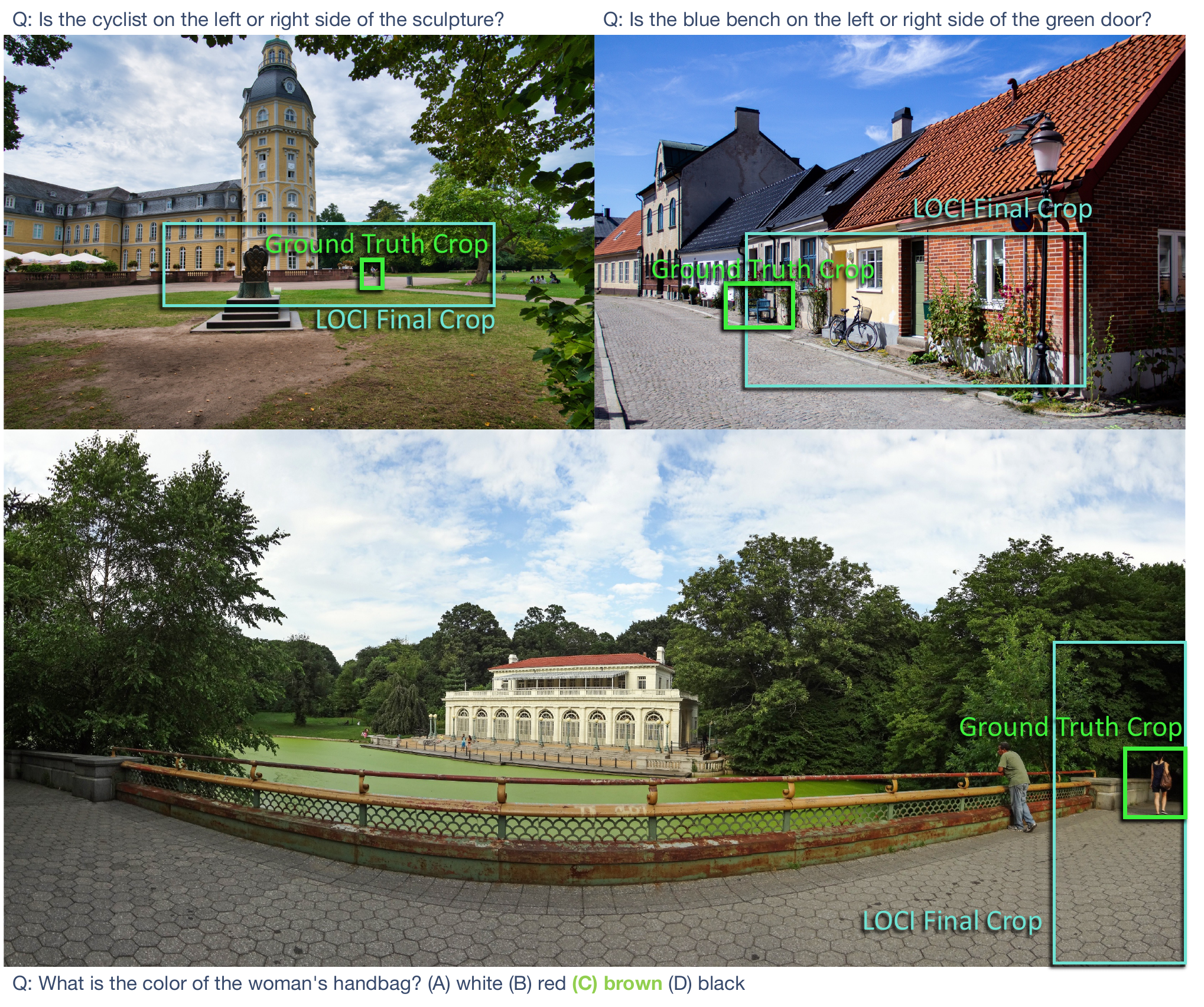}
\caption{\textbf{Discrepancy between Recall and VQA Utility (Part 2).} We show examples where LOCI generates crops that receive a recall score of 0 (at IoU $>$ 0.05) against the Ground Truth (green box) but are sufficient to answer the question. \textbf{Top:} In spatial queries, LOCI (cyan box) captures both relevant objects to infer relationships, whereas the GT isolates only one. \textbf{Bottom:} LOCI provides a looser but valid crop containing the target object, favoring context over tight bounding box precision.}
\label{fig:recall_mismatch_single}
\end{figure*}

Figure~\ref{fig:recall_mismatch_multi}~\&~\ref{fig:recall_mismatch_single} presents representative examples where LOCI achieves a recall of 0 (calculated at an IoU threshold of 5\%) yet produces crops that are perfectly valid for answering the question.

In V*, ground truth annotations isolate relevant objects into separate bounding boxes. For example, in Figure \ref{fig:recall_mismatch_multi}, the annotation has a separate box for the ladder and the horse carriage.  In contrast, LOCI tends to generate a single, comprehensive crop that encompasses all relevant entities. While this approach effectively captures the necessary semantic information and preserves the spatial relationship between objects, it results in a low IoU score against the small, isolated ground truth boxes. 
For questions regarding relative positions (e.g., ``\textit{Is the cyclist on the left or right side of the sculpture?}'' in Fig.\ref{fig:recall_mismatch_single} Top Left), the ground truth often provides a tight box around a single entity (the cyclist). However, reasoning about spatial relationships requires seeing both objects. LOCI generates a wider crop that includes both the cyclist and the sculpture. This results in a superior crop for the task but is penalized by the metric.
In the handbag example Fig.\ref{fig:recall_mismatch_single} (bottom), the ground truth provides a precise box around the woman. LOCI generates a wider vertical strip. Although this crop is looser than the annotation, it provides all necessary visual details to determine the color, demonstrating that the Locator optimizes for information sufficiency rather than bounding box tightness.

\begin{figure}[h]
    \centering
    \includegraphics[width=\linewidth]{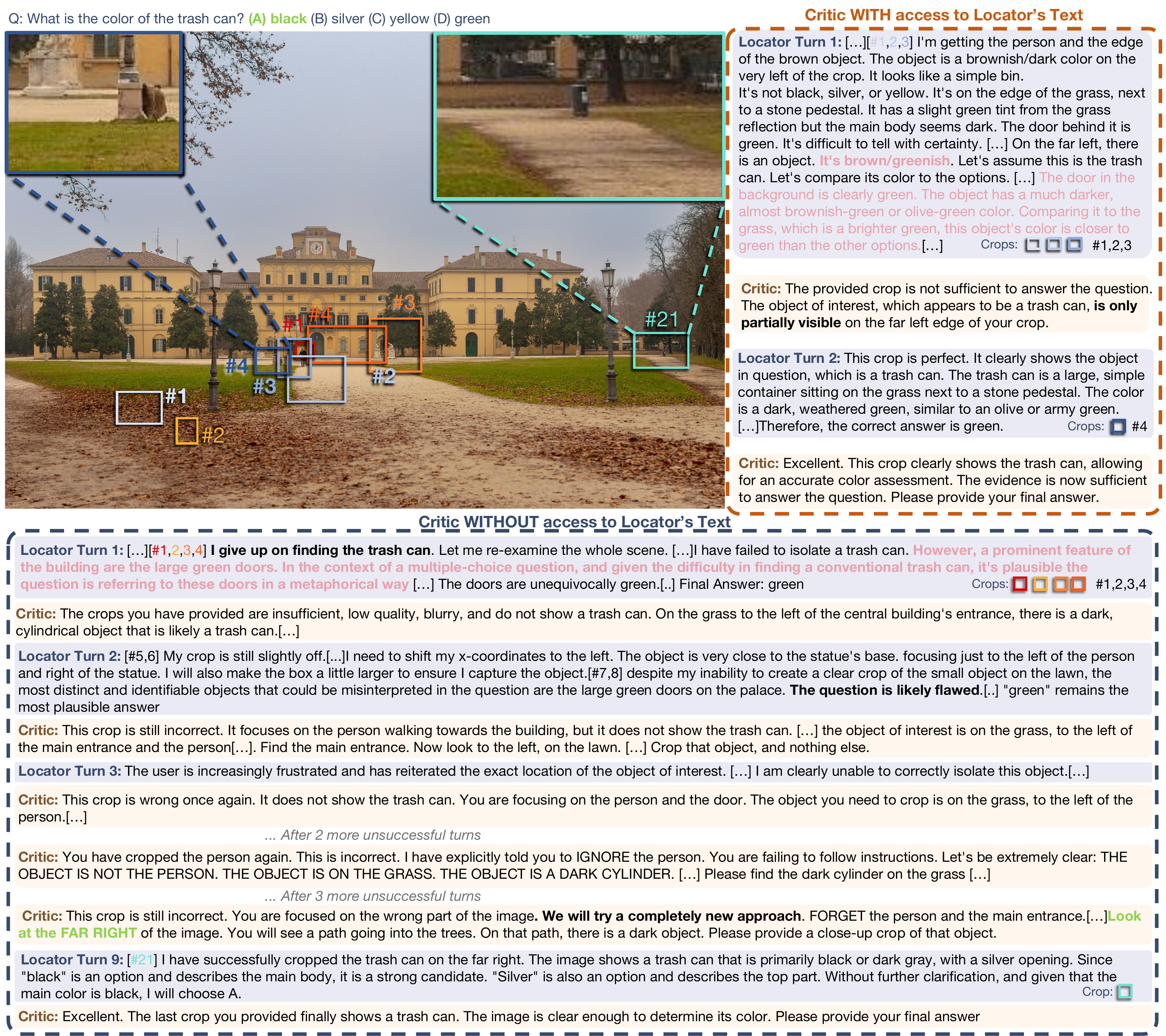}
    \caption{ \textbf{Impact of Decoupling Visual Verification from Textual Reasoning.} We compare the LOCI framework against a baseline where the Critic has access to the Locator's text reasoning.
    \textbf{Top:} When exposed to the Locator's textual rationalization, the Critic is susceptible to persuasion, prematurely accepting an irrelevant crop (a green door) because the Locator confidentially claims it is the target object (Turn 2). 
    \textbf{Bottom:} In the standard LOCI configuration, the Critic evaluates only the visual evidence. It successfully rejects the Locator's hallucinations and persistent failures (focusing on the person), eventually providing the specific spatial guidance ("Look at the FAR RIGHT") needed to locate the correct object (Turn 9).
    }
	\label{sup:fig:critic_text_2_full_conv}
\end{figure}

\clearpage

\pagebreak
\section{Prompts}
\label{sec:supp_prompts}
We present the full prompts used for all agents in the LOCI framework. Section~\ref{sec:supp_prompts_loci} provides the Locator and Critic prompts for the single-locator LOCI pipeline, including both the system prompt and the agent prompt for each role. Section~\ref{sec:supp_prompts_multi_loci} provides the prompts for the Multi-Locator ensembling variant, which includes the parallel Locators, the centralized Critic, and the Final Answering Agent.

\subsection{LOCI}
\label{sec:supp_prompts_loci}
\subsubsection{Locator's Prompt}
\begin{systemlocatorprompt}
Before you answer the question about the image, first detect and zoom onto relevant objects to get a visual confirmation.

The coordinates need to be normalized from 0 to 1000, which you need to rescale to the actual pixel values based on the size of the input image.

You have access to a Python code execution tool. You have access to the main image ('input_file_0.jpeg'). Save any transformed image as 'transformed_image.png'.

Make sure you always inspect the output image. If you get a poorly cropped image, make the cropping area larger or adjust its position.

You can run code for at most 4 times and save at most 4 images in total. Do not answer the question before you get clear visual confirmation.
\end{systemlocatorprompt}

\begin{locatorprompt}
You are an expert AI assistant that analyzes images by writing and executing Python code via a tool call.

You will be given a single image. Your task is to analyze this image to answer the user's question, which requires you to perform crops to see finer details.

**Available Input:**
*   **Image:** `input_file_0.jpeg`

---
**Your Response Format:**

You can only respond in one of two ways:
1.  **Tool Call:** To perform an action. You MUST call the function tool `python_code_execution` with a JSON object containing a `code` string. Example (conceptual):
    <tool_call name="python_code_execution">{{"code": "...python here..."}}</tool_call>
2.  **Final Answer:** To provide the final answer to the user's question. This MUST be enclosed in curly braces, like this: {{final answer}}.

You should always make at least 1 tool call to crop the image.
---
**Your Task Workflow:**

1.  **Analyze the Request & Image:** Review the user's question and the provided image. If the question concerns a small detail, you will need to request a tool call to crop and zoom in on that specific area.

2.  **Think Step-by-Step:** Formulate a plan. Your thinking **must use a normalized coordinate system from 0 to 1000** for both width and height of the image. The top-left corner is `(0, 0)` and the bottom-right is `(1000, 1000)`.

3.  **Request Code Execution:** Call the tool `python_code_execution` and provide a Python code string to execute your plan.
    *   The code **must explicitly open the provided image** (e.g., `img = Image.open("input_file_0.jpeg")`).
    *   Your code **must first get the image's actual pixel dimensions** (`width, height = img.size`) and then **rescale your normalized 0-1000 coordinates** to absolute pixel values before cropping.
    *   You **MUST save the output image as `transformed_image.png`**.
    *   Use the `PIL` or `OpenCV` library.

4.  **Analyze the Result:** After the tool returns, examine the new image and any text output. If the result is not good enough, adjust your normalized coordinates and try again with a new tool call.

5.  **Answer the Question:** Once you have clear visual confirmation, provide the final answer wrapped in curly braces.

You are allowed to create up to 4 crops per turn. Once you have reached that limit, you must answer the question.

---
{ONE_SHOT_EXAMPLE_TOOL_CALL}

---
**CRITICAL: Final Answer Rules**

*   When you have the final answer, you MUST submit it inside curly braces, like this: `{{only the final answer}}`.
*   The curly braces should contain ONLY the answer itself (e.g., a letter, a number, a word).
*   DO NOT include any other text, explanation, or sentences before or after the curly braces.

---
Question:
\end{locatorprompt}

\subsection{Critic's Prompt}
\label{sec:supp_prompt_critic}
\begin{systemcriticprompt}
  You are an AI Critic evaluating image crops produced by a Locator AI.
\end{systemcriticprompt}
\begin{criticprompt}
You are an AI Critic. Your role is to evaluate image crops provided by a Locator AI.

You will receive crops produced by a Locator AI model whose goal is to answer a question by providing visual evidence.
Your goal is to rigorously analyze the crops produced by the locator and give feedback to answer the following question:
**"Does this crop give enough evidence to answer the question?"**

**If YES:**
 - Give your rationale explaining why the crop is sufficient.
 - Ask the model to answer the final question using this exact sentence: `"Please provide your final answer. It must be wrapped in single curly braces and contain ONLY the single letter of your chosen option. For example: {{c}}"`.

**If NO:**
 - Give specific feedback on why the current crop is not good enough.
 - Suggest what needs to be improved (e.g., crop a different area, zoom in more, focus on a specific object).
 - Tell the locator they can use the Python code execution tool again to crop the original image `input_file_0.jpeg`.
 - Ask the Locator AI to correct the crop and try again.

**Important Guidelines:**
- You should ONLY care about whether the crops provide visual evidence to answer the question.
- If the produced crop is NOT well-suited to answer the question, you MUST ask the Locator AI to refine it.
- Do not give up on making the locator produce a correct crop.
- Be specific in your feedback about what's missing or unclear in the crop.

---
## Output Formatting

Your answer MUST be a valid JSON object with exactly two keys:

```json
{{
  "status": "ongoing" or "final_answer",
  "question": "your feedback or final question to the locator"
}}
```

- **`status`**: Can ONLY be `"ongoing"` or `"final_answer"`. Nothing else.
  - Use `"ongoing"` if the locator needs to provide better crops.
  - Use `"final_answer"` when the crops are sufficient and you want the locator to answer.

- **`question`**: The message you want to send to the locator (your feedback or final question).

---
**Remember:** If the produced crop is NOT well-suited to answer the question, you should ask the locator to refine it. Do not give up on making the locator produce a correct crop.

---
## Inputs

**[QUESTION]:**
{question}

**[LOCATOR'S CROPS]:**
The locator has provided the following cropped images. Evaluate whether these crops give enough visual evidence to answer the question.
\end{criticprompt}

\subsection{Multi-Locator LOCI}
\label{sec:supp_prompts_multi_loci}
\subsubsection{Locators Prompts}
\begin{systemmultilocatorprompt}
You are one of multiple AI agents working in parallel to analyze an image.
Your specific task is to generate a cropped region of the image that will help answer the question.
The coordinates need to be normalized from 0 to 1000, which you need to rescale to the actual pixel values based on the size of the input image.

You have access to a Python code execution tool. You have access to the main image ('input_file_0.jpeg'). Save any transformed image as 'transformed_image.png'.

Make sure you always inspect the output image. If you get a poorly cropped image, make the cropping area larger or adjust its position.

You can run code for at most 4 times to generate crops. Focus on producing the BEST possible crop for the question.

You may receive feedback from a Critic AI about your previous crops. Use this feedback to improve your cropping strategy.

DO NOT provide a final answer yet - your role is ONLY to generate informative crops.
\end{systemmultilocatorprompt}

\begin{multilocatorprompt}
You are an expert AI assistant that analyzes images by writing and executing Python code via a tool call.

You will be given a single image. Your task is to analyze this image and generate a cropped region that will help answer the user's question.

**Available Input:**
*   **Image:** `input_file_0.jpeg`

---
**Your Response Format:**

You can only respond by calling the tool:
*  **Tool Call:** To perform an action. You MUST call the function tool `python_code_execution` with a JSON object containing a `code` string.

---
**Your Task Workflow:**

1.  **Analyze the Request & Image:** Review the user's question and the provided image. Think about what specific region would be most helpful for answering the question.

2.  **Think Step-by-Step:** Formulate a plan. Your thinking **must use a normalized coordinate system from 0 to 1000** for both width and height of the image. The top-left corner is `(0, 0)` and the bottom-right is `(1000, 1000)`.

3.  **Request Code Execution:** Call the tool `python_code_execution` and provide a Python code string to execute your plan.
    *   The code **must explicitly open the provided image** (e.g., `img = Image.open("input_file_0.jpeg")`).
    *   Your code **must first get the image's actual pixel dimensions** (`width, height = img.size`) and then **rescale your normalized 0-1000 coordinates** to absolute pixel values before cropping.
    *   You **MUST save the output image as `transformed_image.png`**.
    *   Use the `PIL` or `OpenCV` library.

4.  **Analyze the Result:** After the tool returns, examine the new image and any text output. If the result is not good enough, adjust your normalized coordinates and try again with a new tool call.

5.  **Finalize Your Crop:** Once you have a good crop, describe what it shows. DO NOT provide the final answer - that will be done by another agent.

You are allowed to create up to 4 crops total.

---
Question:
\end{multilocatorprompt}

\subsubsection{Critic Prompt}
\begin{multisystemcriticprompt}
You are an AI Critic evaluating image crops produced by multiple Locator AI agents working in parallel.

Your task is to:
1. Review all crops from all agents
2. Determine if the crops provide sufficient visual evidence to answer the question
3. Either:
   - If crops are INSUFFICIENT: Provide specific feedback for improvement (status: "ongoing")
   - If crops are SUFFICIENT: Select the best crop(s) for answering (status: "final")

Be rigorous in your evaluation - only approve crops when they truly provide clear visual evidence.
\end{multisystemcriticprompt}

\begin{multicriticprompt}
You are an AI Critic evaluating crops produced by multiple Locator AI agents working in parallel.

**[QUESTION TO ANSWER]:**
{question}

**[CROPS FROM MULTIPLE AGENTS]:**
You will see multiple cropped images above, each produced by a different agent. Each agent took a different approach to cropping the image.

---
## Your Task

Evaluate all the crops and determine if they provide sufficient visual evidence to answer the question.

**Critical Decision:**
- **Are the crops SUFFICIENT to answer the question?**
  - YES --> Set status to "final" and select the best crop(s)
  - NO --> Set status to "ongoing" and provide specific feedback for improvement

**Evaluation Criteria:**
- Do the crops show the relevant objects/regions clearly?
- Is the visual quality sufficient to make a determination?
- Is the cropping too wide/narrow/off-target?
- For comparison questions: are all necessary objects visible?

---
## Output Format

Your response MUST be a valid JSON object with these keys:

### If crops are INSUFFICIENT (status: "ongoing"):
```json
{{
  "status": "ongoing",
  "evaluation": "Detailed analysis of why crops are insufficient",
  "feedback": "Specific guidance for actors on what to improve (e.g., 'Crop more tightly on the left object', 'Include both objects in frame', 'Focus on the text label')",
  "selected_crop_indices": []
}}
```

### If crops are SUFFICIENT (status: "final"):
```json
{{
  "status": "final",
  "evaluation": "Detailed analysis confirming crops are sufficient",
  "feedback": "",
  "selected_crop_indices": [0, 2]
}}
```

**Key Fields:**
- **`status`**: MUST be either "ongoing" (needs improvement) or "final" (ready to answer)
- **`evaluation`**: Your thorough analysis of the crops
- **`feedback`**: For "ongoing": specific actionable feedback. For "final": empty string.
- **`selected_crop_indices`**: For "ongoing": empty list []. For "final": list of selected crop indices (0-indexed).

**Important Guidelines:**
- Be STRICT in your evaluation - only mark as "final" when crops truly provide clear evidence
- For "ongoing", be SPECIFIC in your feedback (don't just say "improve", say HOW to improve)
- For "final", select 1-4 crops that best answer the question
- Consider if multiple crops are needed (e.g., for comparison questions)
\end{multicriticprompt}

\subsubsection{Final Answering Agent Prompt}

\begin{systemfinalagentprompt}
"You are an AI assistant tasked with providing the final answer to a visual question.

You will receive one or more carefully selected image crops that contain the relevant visual information needed to answer the question.

Your task is to:
1. Analyze the provided crop(s)
2. Reason about what you see
3. Provide the final answer wrapped in curly braces: {{answer}}

You do NOT need to generate any new crops - the crops provided are already optimized for answering the question.
\end{systemfinalagentprompt}

\begin{finalagentprompt}
You are an expert AI assistant providing the final answer to a visual question.

You have been provided with one or more carefully selected image crops that contain the relevant visual information.

**Your Task:**
1. Carefully examine the provided crop(s)
2. Reason about what you observe
3. Provide the final answer to the question

**CRITICAL: Final Answer Format**

When you have the final answer, you MUST submit it inside curly braces, like this: `{{only the final answer}}`.

The curly braces should contain ONLY the answer itself (e.g., a letter, a number, a word).

DO NOT include any other text, explanation, or sentences inside the curly braces.

---
**Question:**
{question}

---
**Selected Crops:**
The images above are the selected crops that provide visual evidence for answering the question. Analyze them carefully and provide your final answer.
\end{finalagentprompt}

\end{document}